\pdfoutput=1

\PassOptionsToPackage{table}{xcolor} 
\documentclass[11pt]{article}

\usepackage[final]{acl}

\usepackage{times}
\usepackage{latexsym}

\usepackage[T1]{fontenc}

\usepackage[utf8]{inputenc}

\usepackage{microtype}

\usepackage{inconsolata}

\usepackage{graphicx}

\usepackage{amsmath}
\usepackage{amssymb}
\usepackage{bbm}
\usepackage{booktabs} 
\usepackage{fontawesome} 
\usepackage{makecell} 
\usepackage{multirow}
\usepackage{array}

\title{Draft-Thinking: Learning Efficient Reasoning in Long Chain-of-Thought LLMs}

\author{
 \textbf{Jie Cao\textsuperscript{1}\Thanks{These authors contributed equally to this work.}},
 \textbf{Tianwei Lin \textsuperscript{1}\footnotemark[1]},
 \textbf{Juekai Lin \textsuperscript{1}},
 \textbf{Zhenxuan Fan \textsuperscript{1}},
 \textbf{Yunqi Cao \textsuperscript{1}},
 \\
 \textbf{Bo Yuan \textsuperscript{1}},
 \textbf{Ziyuan Zhao \textsuperscript{2}},
 \textbf{Rolan Yan \textsuperscript{2}},
 \textbf{Wenqiao Zhang\Thanks{Corresponding author} \textsuperscript{1}},
 \textbf{Siliang Tang \textsuperscript{1}}
\\
\\
 \textsuperscript{1} Zhejiang University,
 \textsuperscript{2} Tencent
\\
}

\begin{document}
\maketitle

\begin{abstract}
Long chain-of-thought~(CoT) reasoning improves the problem-solving capability of large reasoning models~(LRMs), but often incurs substantial inference cost and overthinking. Existing efficiency methods primarily rely on prompting, post-hoc compression, or length-regularized optimization; these approaches can be sensitive to model scale, depend on external estimators, or under-allocate reasoning to difficult instances. We propose \textbf{Draft-Thinking}, a lightweight adaptation framework that learns a \textit{model-adapted draft CoT}: a compact, self-contained working derivation that directly supports the final answer. We identify a \textbf{Cross-Scale Draft Mismatch}: concise traces generated by a strong teacher can be too sparse for a smaller target model. To address this mismatch, \textbf{Progressive Draft Adaptation} first transfers a concise reasoning-form prior through supervised fine-tuning (SFT) and then applies two-stage reinforcement learning (RL) with progressively harder data and relaxed response budgets to calibrate the reasoning granularity to the target model. 
The method requires neither post-hoc trajectory compression nor an explicit length-penalty reward. 
At inference time, draft, adaptive, and step-by-step prompts provide efficiency-oriented, balanced, and detailed reasoning modes, respectively. The step-by-step mode largely preserves the original model's accuracy while using fewer tokens. With only 847 training problems and 120 RL optimization steps, Draft-Thinking substantially reduces reasoning cost across six mathematical and scientific reasoning benchmarks. On MATH500, its draft mode reduces the average response length from 5,668 to 986 tokens, an 82.6\% reduction, while achieving 90.6\% accuracy compared with 93.0\% for the original long CoT model.

\end{abstract}
\section{Introduction}

Reinforcement learning over long chain-of-thought~(CoT) trajectories has become a mainstream approach for developing reasoning-oriented large language models~(LRMs) with strong problem-solving capabilities~\cite{comanici2025gemini,openai_gpt5_systemcard_2025}, as exemplified by DeepSeek-R1~\cite{guo2025deepseekr1} and Qwen3~\cite{yang2025qwen3technicalreport}. However, these models often default to long, high-budget reasoning trajectories even when substantially shorter derivations would suffice. Such trajectories may contain repeated problem restatements, redundant verification, backtracking, and unproductive alternative branches, increasing inference latency and computational cost and, in some cases, degrading answer quality. This behavior is commonly referred to as \textit{overthinking}~\cite{chen_do_not_2025,cuadron_danger_2025}.

Existing approaches to improving LRM reasoning efficiency can be broadly grouped into three categories. First, prompt-based methods, such as chain-of-draft~(CoD) and sketch-based reasoning, instruct models to produce concise reasoning traces without additional training~\cite{xu_chain_of_draft_2025,aytes_sketch_of_thought_2025}. Although lightweight, their effectiveness is sensitive to prompt design and model scale, and direct zero-shot application can substantially reduce accuracy on small- and medium-sized reasoning models~\cite{xu_chain_of_draft_2025}. Second, compression- and pruning-based methods reduce the reasoning budget by selecting or removing tokens and steps from a reasoning trajectory~\cite{xia_tokenskip_2025,yuan_not_all_tokens_2025}. Their effectiveness therefore depends on an additional compressor, selector, or importance estimator. Third, length-regularized training methods directly discourage verbose generation through length-related rewards or progressively tighter output constraints~\cite{hou_thinkprune_2025,luo_o1-pruner_2025,song_ConciseR_2025}. While effective at reducing token usage, a global preference for shorter responses may under-allocate reasoning to difficult instances whose solutions require more explicit intermediate transitions. These limitations motivate a different question: can a reasoning model learn to generate a concise yet sufficient reasoning trace whose granularity is adapted to its own capability?

\begin{figure*}[t]
  \includegraphics[width=1\textwidth]{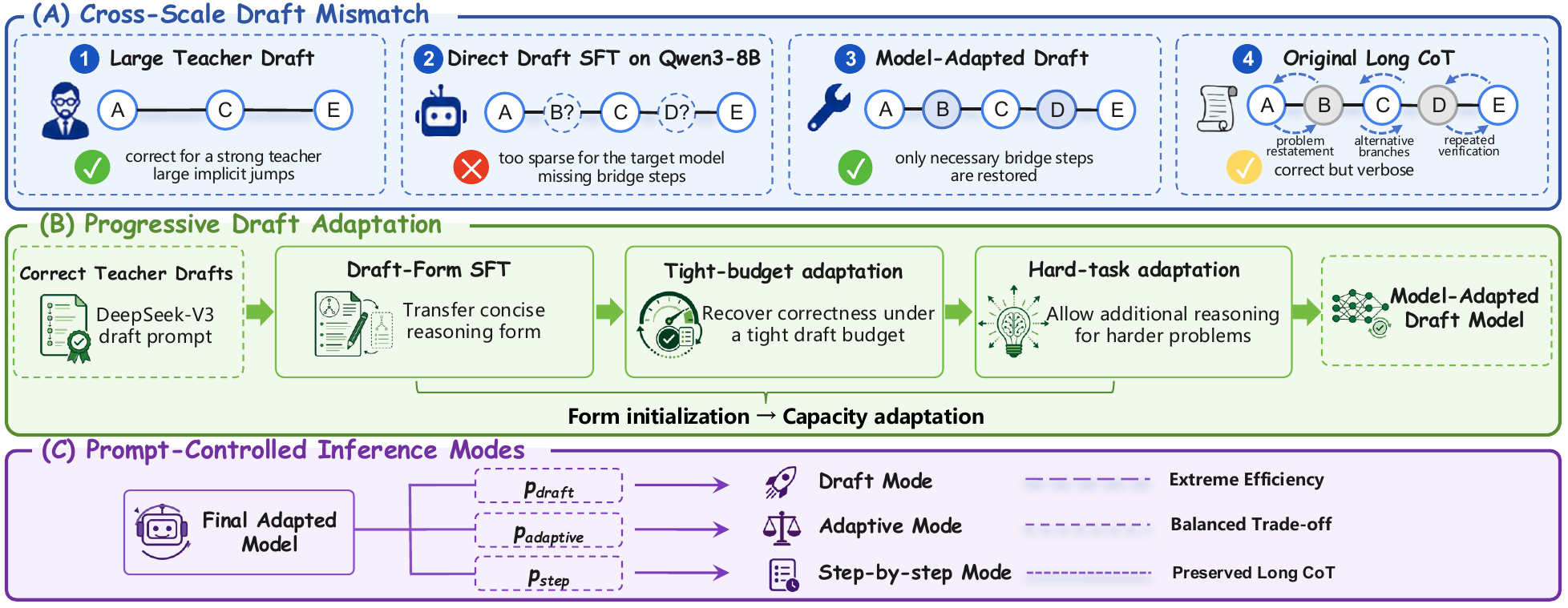}
  \caption{
Overview of Draft-Thinking.
\textbf{(A)} Cross-Scale Draft Mismatch occurs when a strong teacher's concise trace omits intermediate transitions required by a smaller target model.
\textbf{(B)} Progressive Draft Adaptation initializes a concise reasoning form through SFT and then calibrates the draft's reasoning granularity through two correctness-driven RL stages: tight-budget adaptation and hard-task adaptation.
\textbf{(C)} The adapted model supports draft, adaptive, and step-by-step prompts as parallel inference modes that prioritize token efficiency, provide a balanced trade-off, and elicit detailed long CoT reasoning, respectively.
}
  \label{fig:draft-thinking}
\end{figure*}

To this end, we propose \textbf{Draft-Thinking}. We define a \textit{draft CoT} as a compact, self-contained derivation that preserves the answer-sufficient reasoning backbone while omitting verbose restatements, repeated verification, and unproductive alternative branches. Draft-Thinking trains the model to generate such concise reasoning traces directly, rather than compressing or pruning a completed trajectory after generation.

A central challenge is that the level of explicitness required for a reliable draft CoT is model-dependent. A strong teacher can bridge substantial reasoning gaps internally and may therefore express a solution with a sparse explicit trace. The same trace, however, may omit intermediate transitions that a smaller model must express to reason reliably. We refer to this discrepancy as the \textbf{Cross-Scale Draft Mismatch}
(Figure~\ref{fig:draft-thinking}(A)). Consequently, teacher-generated drafts are not universally optimal supervision targets: direct imitation may produce concise traces that omit the explicit reasoning required by the target model, particularly on difficult problems.

To address this mismatch, we introduce \textbf{Progressive Draft Adaptation}, as illustrated in Figure~\ref{fig:draft-thinking}(B). We first perform draft-form supervised fine-tuning~(SFT) on correct draft-style solutions generated by a stronger teacher, initializing the target model with a concise reasoning structure. We then conduct two stages of correctness-driven reinforcement learning~(RL) with progressively harder data and relaxed response budgets. \textbf{Tight-budget adaptation} restores problem-solving accuracy while preserving a compact draft form, whereas \textbf{Hard-task adaptation} permits the additional intermediate transitions required by harder problems. This process yields a \textit{model-adapted draft CoT} that remains substantially shorter than the original long CoT while retaining sufficient explicit reasoning for the target model.

Importantly, Draft-Thinking does not collapse the model into uniformly short responses. As shown in Figure~\ref{fig:draft-thinking}(C), the same adapted model supports three parallel prompt-controlled modes: draft mode prioritizes low-budget reasoning, step-by-step mode elicits detailed long CoT reasoning, and adaptive mode selects an appropriate level of reasoning detail for each instance. Empirically, step-by-step mode largely preserves or improves the original model's accuracy while using fewer tokens. Thus, Draft-Thinking introduces an efficient draft mode without sacrificing access to more expansive reasoning.

Draft-Thinking also requires modest training resources. In our Qwen3-8B setting, the complete pipeline uses only 847 training problems and 120 RL optimization steps, and student-model training finishes in approximately 24 hours on six 48GB NVIDIA L20 GPUs. By adapting the reasoning behavior already present in the base LRM, Draft-Thinking provides a low-cost approach to activating efficient reasoning in existing long CoT models.

Experiments on Qwen3-8B and Qwen3-4B demonstrate substantial reductions in reasoning cost across mathematical and scientific tasks. On MATH500, the Qwen3-8B draft mode reduces the average response length from 5,668 to 986 tokens, an 82.6\% reduction, while achieving 90.6\% accuracy compared with 93.0\% for the original long CoT model. Across all six benchmarks evaluated with Qwen3-8B, Draft-Thinking consistently improves token efficiency.

In summary, our main contributions are:

\noindent \textbf{(i)} We formulate \textbf{model-adapted draft reasoning} as generating a compact, self-contained derivation whose level of explicitness is sufficient for the target model, and identify the \textbf{Cross-Scale Draft Mismatch}: a draft that is sufficient for a strong teacher may be too sparse for a smaller model.

\noindent \textbf{(ii)} We propose \textbf{Progressive Draft Adaptation}, which combines draft-form initialization through SFT with two-stage correctness-driven RL to calibrate draft granularity to the target model. The method requires neither post-hoc trajectory compression nor an explicit length-penalty reward.

\noindent \textbf{(iii)} Experiments across six mathematical and scientific reasoning benchmarks consistently show that Draft-Thinking substantially improves token efficiency while preserving access to detailed long CoT reasoning. These gains are achieved using only 847 training problems and 120 RL optimization steps, demonstrating a practical, low-cost adaptation pathway for existing LRMs.
\section{Related Work}

Recent research has revealed that LRMs suffer from an overthinking issue 
\citep{chen_do_not_2025, cuadron_danger_2025, yang_tops_2025}, where models with long CoT reasoning often expend unnecessary computational resources on redundant solutions that contribute minimally to final outcomes and may even degrade model performance.

\noindent\textbf{Prompt-based Efficient Reasoning.} Prompt-based techniques \citep{xu_chain_of_draft_2025,aytes_sketch_of_thought_2025} improve reasoning efficiency through well-designed prompts with a few-shot setting. CoD \cite{xu_chain_of_draft_2025} utilizes the simple instruction "Think step by step, but only keep a minimum draft for each thinking step, with 5 words at most." significantly reduces token usage for GPT-4o and Claude 3.5 Sonnet. Sketch-of-Thought \cite{aytes_sketch_of_thought_2025} routes each instance to the most appropriate template among three cognitive science-inspired prompts.

\noindent\textbf{Compression of Chain-of-Thought.} C3oT \cite{kang_c3ot_2024} employs GPT-4 as a compressor to distill longer CoTs into shorter versions while preserving key information. It then trains LLMs on both longer and shorter CoT to learn their relationships and finally performs inference using the shorter CoT to achieve efficient reasoning.
TokenSkip \cite{xia_tokenskip_2025} and CTS \cite{yuan_not_all_tokens_2025} compress original CoTs into shorter versions by applying different compression ratios according to token semantic importance, and then perform inference under specific compression ratios to achieve efficient reasoning. 
TALE \cite{han_token-budget-aware_2025} propose a token-budget-aware LLM reasoning framework that adjusts the number of reasoning tokens based on the complexity of each problem.

\noindent\textbf{Length-Regularized Reinforcement Learning.} THINKPRUNE \cite{hou_thinkprune_2025} applies an iterative pruning strategy in RL training with an increasingly stringent token limit to reduce the reasoning length of long CoT LLMs. O1-Pruner \cite{luo_o1-pruner_2025}, Kimi \cite{team_kimi_2025}, and ConciseR \cite{song_ConciseR_2025} incorporate length-based penalties into the reward function for RL training, incentivizing the model to produce more concise reasoning while maintaining accuracy. LC-R1 \cite{cheng_LCR1_2025} incorporates a length reward into Group Relative Policy Optimization (GRPO), and a compress model to remove the invalid portion of the thinking process.
\section{Methodology}

\subsection{Model-Adapted Draft Reasoning}
\label{sec:formulation}

Let $M_{\theta}$ denote an LRM with long CoT capability. Given a question $\boldsymbol{q}$ and a reasoning prompt $\boldsymbol{p}$, the model generates
\begin{equation}
    \boldsymbol{o}
    =
    \boldsymbol{r} \oplus \boldsymbol{a}
    \sim
    M_{\theta}(\cdot \mid \boldsymbol{p}, \boldsymbol{q}),
\end{equation}
where $\boldsymbol{r}$ is the reasoning trace, $\boldsymbol{a}$ is the final answer, and $\oplus$ denotes concatenation. Under the step-by-step prompt $\boldsymbol{p}_{\text{step}}$, the model typically produces a detailed reasoning trace $\boldsymbol{r}_{\text{step}}$.

We define a \textit{model-adapted draft CoT} as a compact and self-contained derivation that preserves an answer-sufficient reasoning backbone at the level of explicitness required by the target model. Draft-Thinking aims to train $M_{\theta}$ such that, under a draft prompt $\boldsymbol{p}_{\text{draft}}$, it generates
\begin{equation}
    \boldsymbol{o}_{\text{draft}}
    =
    \boldsymbol{r}_{\text{draft}} \oplus \boldsymbol{a}
    \sim
    M_{\theta}(\cdot \mid \boldsymbol{p}_{\text{draft}}, \boldsymbol{q}),
\end{equation}
where $\boldsymbol{r}_{\text{draft}}$ is substantially shorter than $\boldsymbol{r}_{\text{step}}$ while remaining sufficient for solving the problem.

\subsection{Progressive Draft Adaptation}
\label{sec:progressive_adaptation}

The level of explicitness required for a reliable draft CoT is
model-dependent. A strong teacher may omit intermediate transitions
that it can resolve internally, whereas the same trace may be too
sparse for a smaller target model. We refer to this discrepancy as the \textbf{Cross-Scale Draft Mismatch}.

As illustrated in Figure~\ref{fig:draft-thinking},
\textbf{Progressive Draft Adaptation} addresses this mismatch through
two phases. Draft-form initialization transfers a concise reasoning structure through SFT, while capacity adaptation uses two stages of correctness-driven RL, namely tight-budget adaptation and hard-task adaptation, to calibrate the draft's reasoning granularity to the target model's capacity.

\subsubsection{Draft-Form Initialization}
\label{sec:draft_stage1}

Prompting alone does not reliably induce accurate draft reasoning, particularly for small- and medium-scale models~\cite{xu_chain_of_draft_2025}. We therefore use a stronger teacher model to initialize the target model toward a concise reasoning form.

Let
\begin{equation}
    \mathcal{D}
    =
    \{(\boldsymbol{q}_i,\boldsymbol{a}_i)\}_{i=1}^{817}
\end{equation}
denote the LIMO dataset~\cite{ye_limo_2025}. We prompt DeepSeek-V3-0324 with the Chunked Symbolism instruction~\cite{aytes_sketch_of_thought_2025} to generate a draft trace $\hat{\boldsymbol{r}}_i$ and an answer $\hat{\boldsymbol{a}}_i$ for each question. Retaining only samples for which $\hat{\boldsymbol{a}}_i=\boldsymbol{a}_i$ yields
\begin{equation}
    \mathcal{D}_{\text{sft}}
    =
    \{(\boldsymbol{q}_i,
       \hat{\boldsymbol{r}}_i,
       \boldsymbol{a}_i)\}_{i=1}^{342}.
\end{equation}

The target model is then fine-tuned under $\boldsymbol{p}_{\text{draft}}$ by minimizing
\begin{equation}
\begin{aligned}
    \mathcal{L}_{\text{sft}}
    =
    -\sum_{(\boldsymbol{q},\boldsymbol{r},\boldsymbol{a})
    \in \mathcal{D}_{\text{sft}}}
    \sum_{t=1}^{|\boldsymbol{y}|}
    \log P_{\theta}
    \left(
        y_t
        \mid
        \boldsymbol{p}_{\text{draft}},
        \boldsymbol{q},
        \boldsymbol{y}_{<t}
    \right),
\end{aligned}
\end{equation}
where $\boldsymbol{y}=\boldsymbol{r}\oplus\boldsymbol{a}$.

This stage initializes the target model with a concise draft-style reasoning form. The resulting policy then serves as the starting point for capacity adaptation, which calibrates its reasoning granularity to the target model.

\subsubsection{Capacity Adaptation via Progressive RL}
\label{sec:draft_rl}

We next adapt the granularity of draft reasoning to the capacity of the target model. Starting RL with a large response budget can cause the model to recover reward by reverting to its original long CoT behavior, even under $\boldsymbol{p}_{\text{draft}}$. We therefore introduce \textbf{Tight-budget adaptation} and \textbf{Hard-task adaptation}, which combine increasingly difficult training data with progressively relaxed maximum response lengths $L_{\max}$.

\begin{itemize}
    \item \textbf{Tight-budget adaptation.}
    We use the remaining 475 LIMO problems $D_{\text{rl}} = \{(\boldsymbol{q}_i, \boldsymbol{a}_i)\}_{i=1}^{475}$ and set
    $L_{\max}=3{,}000$. This stage restores problem-solving
    accuracy while maintaining a compact draft form.

    \item \textbf{Hard-task adaptation.}
    We subsequently train on the 30 AIME2024 problems with
    $L_{\max}=6{,}000$. The relaxed upper bound allows the
    model to introduce additional intermediate transitions
    when they are required by more difficult problems.
\end{itemize}

Increasing $L_{\max}$ enlarges the feasible reasoning space but does not prescribe a target response length. In particular, no explicit response-length term is included in the reward. The optimization is instead driven by task correctness, allowing the model to determine how much explicit reasoning it requires within the permitted budget.

Both stages use GRPO~\cite{shao_deepseekmath_2024}. For each pair $(\boldsymbol{q},\boldsymbol{a})$, we sample a group of outputs
\begin{equation}
    \{\boldsymbol{o}_i\}_{i=1}^{G}
    \sim
    \pi_{\theta_{\text{old}}}
    \left(
        \cdot
        \mid
        \boldsymbol{p}_{\text{draft}},
        \boldsymbol{q}
    \right)
\end{equation}
under the stage-specific response limit. Let $\hat{A}_i$ denote the group-relative advantage computed from the answer-based task reward. The policy is optimized using
\begin{multline}
    \mathcal{J}_{\text{GRPO}}(\theta)
    =
    \mathbb{E}
    \Bigg[
    \frac{1}{G}
    \sum_{i=1}^{G}
    \frac{1}{|\boldsymbol{o}_i|}
    \sum_{t=1}^{|\boldsymbol{o}_i|}
    \min
    \Big(
        \rho_{i,t}(\theta)\hat{A}_i, \\
        \operatorname{clip}
        \big(
            \rho_{i,t}(\theta),
            1-\epsilon,
            1+\epsilon
        \big)
        \hat{A}_i
    \Big)
    \Bigg],
\end{multline}
where
\begin{equation}
    \rho_{i,t}(\theta)
    =
    \frac{
        \pi_{\theta}
        (o_{i,t}
        \mid
        \boldsymbol{p}_{\text{draft}},
        \boldsymbol{q},
        \boldsymbol{o}_{i,<t})
    }{
        \pi_{\theta_{\text{old}}}
        (o_{i,t}
        \mid
        \boldsymbol{p}_{\text{draft}},
        \boldsymbol{q},
        \boldsymbol{o}_{i,<t})
    }.
\end{equation}

Through these two stages, the teacher-derived draft form is progressively adapted into a model-specific draft that retains the compact yet sufficient reasoning steps required by the target model, while remaining substantially shorter than its original long CoT. 

\subsection{Prompt-Controlled Inference Modes}
\label{sec:inference}

At inference time, the adapted model supports three prompt-controlled reasoning modes:

\begin{itemize}
    \item \textbf{Draft mode} uses
    $\boldsymbol{p}_{\text{draft}}$ to invoke the learned
    low-budget draft reasoning behavior.

    \item \textbf{Step-by-step mode} uses
    $\boldsymbol{p}_{\text{step}}$ to elicit more detailed
    long CoT reasoning.

    \item \textbf{Adaptive mode} uses
    $\boldsymbol{p}_{\text{adaptive}}$ to instruct the model to select an appropriate reasoning granularity.
\end{itemize}

The same adapted model supports all three modes independently.
Draft mode prioritizes token efficiency, step-by-step mode provides a more accuracy-oriented operating point, and adaptive mode offers an intermediate trade-off. The prompt templates are provided in Table~\ref{tab:prompts}.
\section{Experiments}

\subsection{Experiment Setup}

\begingroup
\setlength{\tabcolsep}{3pt}
\begin{table*}[t]
  \centering
  \begin{tabular}{lcccccccccccc}
    \Xhline{1.5pt}
    \multirow[m]{2}{*}{\textbf{Method}} &  \multicolumn{3}{c}{\textbf{MATH500}} & \multicolumn{3}{c}{\textbf{AIME2025}} & \multicolumn{3}{c}{\textbf{LiveMathBench}} & \multicolumn{3}{c}{\textbf{OlympiadBench}} \\
    
    &Acc&Tokens&EFF &Acc&Tokens&EFF &Acc&Tokens&EFF &Acc&Tokens&EFF \\
    \hline
    \multicolumn{13}{c}{\texttt{Qwen3-8B}} \\
    \hline
    Original &93.00&5668&1.64  &64.38&18034&0.36 &32.17&17493&0.18 &64.44&11887&0.54 \\ %

    SFT-only &91.00&4364&2.08 &49.58&13129&0.37 &26.17&13160&0.19 &57.33&10424&0.55 \\  
    
    GRPO-only &93.60&3074&3.04 &53.96&11720&0.46 &28.67&11043&0.26 &65.48&6902&0.94 \\ 
    
    ThinkPrune &94.40&2615&\underline{3.61}  &48.13&9781&\underline{0.49} &28.67&9333&\underline{0.31} &65.63&5809&\underline{1.13} \\ 
    
    \hline
    \rowcolor[HTML]{F2F2F2}
    Draft-Thinking &90.60&986&\textbf{9.18} &45.42&6527&\textbf{0.69} &22.33&6476 &\textbf{0.35} &61.19&3433&\textbf{1.78}\\ 
    \Xhline{1.5pt}
  \end{tabular}
  \caption{\label{tab:results}
    Controlled comparison of draft-mode reasoning efficiency on four mathematical reasoning benchmarks using Qwen3-8B. The highest and second-highest EFF values are shown in \textbf{bold} and \underline{underlined}, respectively. Draft-Thinking is evaluated under the draft prompt $\boldsymbol{p}_{\mathrm{draft}}$.
  }
\end{table*}
\endgroup

\begingroup
\setlength{\tabcolsep}{3pt}
\begin{table}[t]
\small
  \centering
  \begin{tabular}{lcccccc}
    \Xhline{1.5pt}
    \multirow[m]{2}{*}{\textbf{Method}} &   \multicolumn{3}{c}{\textbf{Minerva}} & \multicolumn{3}{c}{\textbf{GPQA-D}} \\
    
    &Acc&Tokens&EFF &Acc&Tokens&EFF \\
    \hline
    \multicolumn{7}{c}{\texttt{Qwen3-8B}} \\
    \hline
    Original &52.21&7640&0.68 &55.05&11554&0.47 \\ %

    SFT-only &47.06&7349&0.64 &37.37&16367&0.23 \\  
    
    GRPO-only  &53.68&3822&1.40 &60.61&6322&0.96 \\ 
    
    ThinkPrune &54.04&2725&\underline{1.98} &60.10&5166&\underline{1.16} \\ 
    \hline
    \rowcolor[HTML]{F2F2F2}
    Draft-Thinking &46.32&960&\textbf{4.82} &54.04&2689&\textbf{2.01} \\ 
    \Xhline{1.5pt}
  \end{tabular}
  \caption{\label{tab:transfer_results}
    Controlled comparison on two transfer benchmarks using Qwen3-8B.
  }
\end{table}
\endgroup

\begingroup
\setlength{\tabcolsep}{0.8pt}
\begin{table*}[t]
\small
  \centering
  \begin{tabular}{llllllllllllll}
    \Xhline{1.5pt}
    \multirow[m]{2}{*}{\textbf{Method}} & \multirow[m]{2}{*}{\textbf{Prompt}} & \multicolumn{2}{c}{\textbf{MATH500}} & \multicolumn{2}{c}{\textbf{AIME2025}} & \multicolumn{2}{c}{\textbf{LiveMathBench}} & \multicolumn{2}{c}{\textbf{OlympiadBench}} & \multicolumn{2}{c}{\textbf{Minerva}} & \multicolumn{2}{c}{\textbf{GPQA-D}} \\
    
    & &Acc/Tokens&EFF &Acc/Tokens&EFF &Acc/Tokens&EFF &Acc/Tokens&EFF &Acc/Tokens&EFF &Acc/Tokens&EFF \\
    \hline
    \multicolumn{14}{c}{\texttt{Qwen3-8B}} \\
    \hline
    Original &step &93.00/5668&1.64  &\underline{64.38}/18034&0.36 &\textbf{32.17}/17493 &0.18 &64.44/11887&0.54 &\underline{52.21}/7640&0.68 &\underline{55.05}/11554&0.47 \\ %

    \rowcolor[HTML]{F2F2F2}
    \multirow[t]{3}{*}{Draft-Thinking} &step &\textbf{93.60}/4271&2.19 &\textbf{64.79}/14112&0.46 &30.83/13980&0.22 &\underline{66.37}/9519&0.69 &\textbf{53.68}/5180&1.03 &\textbf{63.13}/6620&0.95 \\ 
         
    \rowcolor[HTML]{F2F2F2}
    &adaptive &\underline{93.20}/2755&\underline{3.38} &56.67/12155&\underline{0.46} &\underline{31.83}/10341&\underline{0.31} &\textbf{69.19}/6849&\underline{1.01} &48.53/2979&\underline{1.62} &52.02/4800&\underline{1.08} \\ 
    
    \rowcolor[HTML]{F2F2F2}
    &draft &90.60/986&\textbf{9.18} &45.42/6527&\textbf{0.69} &22.33/6476&\textbf{0.35} &61.19/3433&\textbf{1.78} &46.32/960&\textbf{4.82} &54.04/2689&\textbf{2.01} \\ 
     
     \Xhline{1.5pt}
  \end{tabular}
  \caption{\label{tab:inference_modes_results}
    Comparison of three prompt-controlled inference modes of the same adapted Qwen3-8B model. The draft, adaptive, and step-by-step rows share identical model parameters and differ only in the inference prompt.
  }
\end{table*}
\endgroup

\begingroup
\setlength{\tabcolsep}{0.5pt}
\begin{table*}[t]
\small
  \centering
  \begin{tabular}{llllllllllllll}
    \Xhline{1.5pt}
    \multirow[m]{2}{*}{\textbf{Method}} & \multirow[m]{2}{*}{\textbf{Prompt}} & \multicolumn{2}{c}{\textbf{MATH500}} & \multicolumn{2}{c}{\textbf{AIME2025}} & \multicolumn{2}{c}{\textbf{LiveMathBench}} & \multicolumn{2}{c}{\textbf{OlympiadBench}} & \multicolumn{2}{c}{\textbf{Minerva}} & \multicolumn{2}{c}{\textbf{GPQA-D}} \\
    & &Acc/Tokens&EFF &Acc/Tokens&EFF &Acc/Tokens&EFF &Acc/Tokens&EFF &Acc/Tokens&EFF &Acc/Tokens&EFF \\
    \hline
    \multicolumn{14}{c}{\texttt{Qwen3-8B}} \\
    \hline
    Original& step &93.00/5668&1.64  &64.38/18034&0.36 &32.17/17493&0.18 &64.44/11887&0.54 &52.21/7640&0.68 &55.05/11554&0.47 \\ %

    Original &draft &95.00/3714&2.55 &65.62/16254&0.40 &27.00/17990&0.15 &65.33/9968&0.65 &51.47/4889&1.05 &57.07/8143&0.70 \\
    
    D-SFT &draft &69.80/1732&4.02 &9.17/3928&0.23 &12.50/3732&0.33 &36.89/3535&1.04 &37.87/549&6.88 &41.92/1948&2.15 \\ 
    
    D-SFT->D-RL6k &draft &93.00/1453&6.39 &48.75/7426&0.65 &26.33/6949&0.37 &63.41/4340&1.46 &49.26/1677&2.93 &57.07/3742&1.52 \\

    D-SFT->D-RL3k &draft &89.40/1032&8.65 &43.96/6100&0.72 &22.33/5002&0.45 &57.78/3785&1.52 &44.49/1133&3.92 &48.99/4504&1.08 \\
    
    \rowcolor[HTML]{F2F2F2}
    Draft-Thinking &draft &90.60/986&9.18 &45.42/6527&0.69 &22.33/6476&0.35 &61.19/3433&1.78 &46.32/960&4.82 &54.04/2689&2.01 \\
    \Xhline{1.5pt}
    
  \end{tabular}
  \caption{\label{tab:ablation_results}
  Ablation study of Draft-Thinking on Qwen3-8B. Original (step) is the unadapted long CoT reference, whereas Original (draft) applies $\boldsymbol{p}_{\mathrm{draft}}$ without training. D-SFT uses draft-form SFT only; D-SFT$\rightarrow$D-RL6k omits tight-budget adaptation and performs all 120 RL steps with $L_{\max}=6{,}000$; D-SFT$\rightarrow$D-RL3k (60 steps) is the intermediate checkpoint after tight-budget adaptation; and Draft-Thinking is the full D-SFT$\rightarrow$D-RL3k (60 steps)$\rightarrow$D-RL6k (60 steps) pipeline.
  }
\end{table*}
\endgroup

\paragraph{Backbone models.}
We use Qwen3-8B~\cite{yang2025qwen3technicalreport} as the primary backbone for controlled baseline comparisons and main experiments. We additionally conduct experiments on Qwen3-4B to examine the effectiveness of Draft-Thinking at a smaller scale, with results reported in the Appendix~\ref{appendix:training pipeline}.

\paragraph{Training data.}
The complete Draft-Thinking pipeline uses 847 training questions. The draft-form SFT dataset contains
$\mathcal{D}_{\mathrm{sft}}
=
\{(\boldsymbol{q}_i,\hat{\boldsymbol{r}}_i,\boldsymbol{a}_i)\}_{i=1}^{342}$.
The first RL stage uses the remaining 475 questions from LIMO~\cite{ye_limo_2025},
denoted as
$\mathcal{D}_{\mathrm{rl}}
=
\{(\boldsymbol{q}_i,\boldsymbol{a}_i)\}_{i=1}^{475}$,
and the second RL stage uses the 30 problems from AIME2024.
The two LIMO subsets are disjoint, as described in
Sections~\ref{sec:draft_stage1}.

\paragraph{Evaluation benchmarks.}
We evaluate the models on six mathematical and scientific reasoning benchmarks.
MATH500~\cite{hendrycks2021math500}, AIME2025,
LiveMathBench v202505~\cite{liu2024livemathbench202505},
and OlympiadBench~\cite{he2024olympiadbench} are used to evaluate mathematical reasoning under diverse levels of difficulty.
We further evaluate transfer beyond the training distribution using
MinervaMath~\cite{lewkowycz2022minervamath},
which contains scientific quantitative problems, and
GPQA-Diamond~\cite{rein2024gpqa},
which evaluates graduate-level scientific reasoning.

\paragraph{Evaluation protocol.}
We report pass@1 accuracy (\textbf{ACC}), the average number of generated response tokens (\textbf{Tokens}), and token efficiency (\textbf{EFF}).
The detailed evaluation protocol is provided in the Appendix~\ref{appendix:evaluation}.

\paragraph{Controlled baselines.}
To isolate the contribution of Draft-Thinking, we implement three baselines with the same Qwen3-8B backbone: \textbf{SFT-only}, \textbf{GRPO-only}~\cite{shao_deepseekmath_2024}, and \textbf{ThinkPrune}~\cite{hou_thinkprune_2025} (see Appendix~\ref{appendix:baselines} for details).


\subsection{Main Experimental Results}

\paragraph{Draft mode provides the strongest token-efficiency operating point.}
Tables~\ref{tab:results} and~\ref{tab:transfer_results} compare the final Draft-Thinking model in its primary draft mode with controlled baselines using the same Qwen3-8B backbone.
Draft-Thinking achieves the highest EFF among the compared methods on all six benchmarks, demonstrating that draft-form initialization and progressive adaptation produce substantially more concise reasoning than SFT-only, GRPO-only, and ThinkPrune.

On MATH500, Draft-Thinking reduces the average response length from 5,668 to 986 tokens, corresponding to an 82.6\% reduction.
Its accuracy changes from 93.0\% to 90.6\%, an absolute decrease of 2.4 percentage points, while EFF increases from 1.64 to 9.18, representing a 5.6$\times$ improvement.
These results show that the model can preserve a substantial portion of its reasoning performance under a much tighter inference budget.

\paragraph{The accuracy--efficiency trade-off becomes more pronounced on the hardest tasks.}
Draft mode is designed as the low-budget operating point of Draft-Thinking rather than the accuracy-maximizing configuration.
On AIME2025 and LiveMathBench, Draft-Thinking continues to achieve the highest EFF, but the reduction in explicit reasoning is accompanied by a larger accuracy decrease.
For example, on AIME2025, Draft-Thinking reduces the average response length by 63.8\% relative to the original model, while accuracy decreases from 64.38\% to 45.42\%.
These results suggest that highly challenging competition problems require more explicit intermediate reasoning than easier mathematical tasks.
We examine more accuracy-oriented operating modes in Table~\ref{tab:inference_modes_results}.

\paragraph{Draft reasoning transfers beyond the mathematical training distribution.}
Table~\ref{tab:transfer_results} shows that the learned draft behavior also transfers to scientific quantitative and scientific reasoning tasks.
On GPQA-Diamond, draft mode reduces the average response length from 11,554 to 2,689 tokens, a 76.7\% reduction, while accuracy decreases by only 1.01 percentage points from 55.05\% to 54.04\%.
Consequently, EFF increases from 0.47 to 2.01.


\subsection{Prompt-Controlled Inference Modes}



Table~\ref{tab:inference_modes_results} compares the three prompt-controlled inference modes of the same adapted Qwen3-8B model.
Draft mode represents the lowest-budget operating point, step-by-step mode evaluates whether draft-oriented training preserves detailed long CoT reasoning, and adaptive mode provides an intermediate accuracy--efficiency trade-off.

\paragraph{Draft-oriented training preserves and regularizes long CoT reasoning.}
Under the step-by-step prompt, the adapted model improves accuracy on five of the six benchmarks and improves EFF on all six benchmarks relative to the original Qwen3-8B model.
This indicates that learning draft reasoning does not collapse the model into uniformly short responses.
Instead, the training procedure also reduces redundancy when the model is prompted to produce detailed long COT reasoning.

On AIME2025, long CoT accuracy increases slightly from 64.38\% to 64.79\%, while the average response length decreases from 18,034 to 14,112 tokens, corresponding to a 21.7\% reduction.
The largest improvement is observed on GPQA-Diamond, where accuracy increases by 8.08 percentage points from 55.05\% to 63.13\%, while response length decreases by 42.7\% from 11,554 to 6,620 tokens.


\paragraph{Adaptive mode offers an intermediate accuracy–efficiency trade-off.}
Across all benchmarks, adaptive mode consistently achieves higher accuracy than draft mode and exceeds the EFF of step-by-step mode.
It therefore recovers part of the accuracy lost under extreme draft compression while remaining substantially more concise than long CoT reasoning.

For example, 
on OlympiadBench, adaptive mode improves accuracy from 66.37\% to 69.19\% relative to step-by-step mode while reducing token usage from 9,519 to 6,849 tokens, a 28.0\% reduction.

\subsection{Ablation study of Draft-Thinking}
\label{sec:ablation}

\begin{figure*}[t]
  \includegraphics[width=1\textwidth]{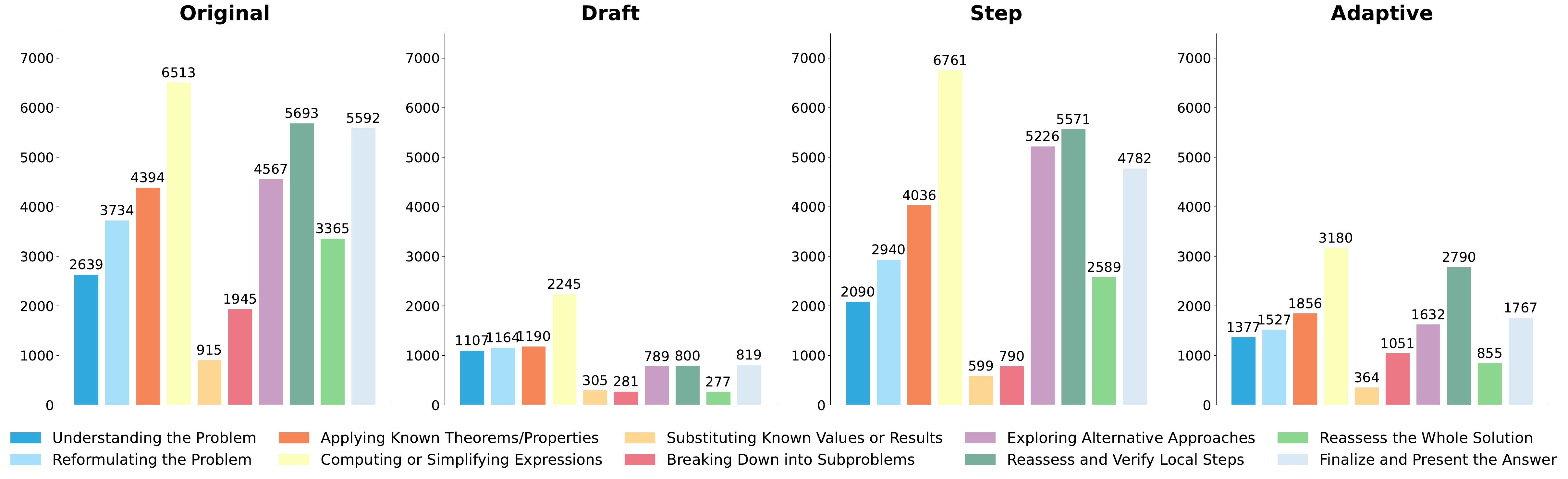}
  \caption{Reasoning behavior comparison between original Qwen3-8B and Draft-Thinking on MATH500. Each bar represents the cumulative number of reasoning steps within a phase category.}
  \label{fig:reasoning_behavior}
\end{figure*}


Table~\ref{tab:ablation_results} isolates the contributions of draft prompting only, draft-form initialization, tight-budget adaptation, and hard-task adaptation.

\paragraph{Draft prompting exposes reasoning redundancy but does not reliably elicit compact reasoning.}
Applying $\boldsymbol{p}_{\mathrm{draft}}$ directly to the unadapted Qwen3-8B improves EFF on five of the six benchmarks, suggesting that its original long CoT traces contain substantial redundancy. However, the prompt-only draft reasoning still generates 16,254 tokens on AIME2025 and 17,990 tokens on LiveMathBench; These results show that prompting alone cannot reliably produce concise and accurate draft reasoning, motivating the need to learn a model-adapted draft behavior.

\paragraph{Draft-form SFT alone exposes the Cross-Scale Draft Mismatch.}
Draft-form SFT sharply shortens reasoning traces but substantially degrades accuracy across all six benchmarks. 
On AIME2025, relative to the Original (draft) baseline, D-SFT reduces the average response length by 75.8\% (16,254$\rightarrow$3,928), while accuracy drops by 56.45 percentage points (65.62\%$\rightarrow$9.17\%). 

These results are consistent with the proposed Cross-Scale Draft Mismatch: teacher-generated drafts establish a concise reasoning-form prior but may omit intermediate transitions required by Qwen3-8B for reliable reasoning. Therefore, D-SFT serves as form initialization rather than complete adaptation, motivating the subsequent capacity-adaptation stages to restore correctness while preserving conciseness.

\paragraph{Tight-budget adaptation stabilizes compact reasoning.}
The first RL stage recovers a substantial portion of the accuracy lost after D-SFT while preserving a low-budget draft form. From D-SFT to D-SFT$\rightarrow$D-RL3k, accuracy improves across all six benchmarks, including from 69.80\% to 89.40\% on MATH500 and from 9.17\% to 43.96\% on AIME2025.

The contribution of this stage is best isolated by comparison with D-SFT$\rightarrow$D-RL6k, which omits tight-budget adaptation while using the same total RL optimization budget. The complete Draft-Thinking pipeline generates fewer tokens on all six benchmarks and achieves higher EFF on five, with an accuracy trade-off of 2.22--4.00 percentage points. These results indicate that the initial 3k stage steers subsequent optimization toward compact reasoning, reducing reliance on longer trajectories during the later 6k stage.

\paragraph{Hard-task adaptation further calibrates draft granularity.}
Compared with the intermediate D-SFT$\rightarrow$D-RL3k model, the complete Draft-Thinking model improves accuracy on five of the six benchmarks. On MATH500, OlympiadBench, MinervaMath, and GPQA-Diamond, these gains are also accompanied by further reductions in response length. For example, on GPQA-Diamond, accuracy increases by 5.05 percentage points while the average response length decreases from 4,504 to 2,689 tokens. These results suggest that hard-task adaptation helps the model use its reasoning budget more selectively, rather than uniformly expanding its reasoning traces.

The pattern differs on AIME2025 and LiveMathBench. On AIME2025, accuracy increases from 43.96\% to 45.42\%, while the average response length rises from 6,100 to 6,527 tokens. Consequently, the hard-task adaptation does not improve EFF on these two challenging benchmarks, suggesting that the current 6k adaptation budget or limited hard-task training set is not yet sufficient to reliably construct model-adapted drafts for the hardest problems. Improving these cases may require a larger feasible response budget or more diverse hard-problem training data.

Overall, the ablation results confirm the complementary roles of the three training stages. Draft-form SFT initializes a concise reasoning form, tight-budget adaptation restores correctness while constraining reasoning expansion, and hard-task adaptation further calibrates the amount of explicit reasoning under more difficult data. Together, the three stages transform teacher-generated drafts into model-adapted draft CoT that better balance correctness and conciseness.

\begingroup
\setlength{\tabcolsep}{3pt}
\begin{table}[t]
\small
    \centering
    \begin{tabular}{lcccc}
    \Xhline{1.5pt}
    \textbf{Metric} & \textbf{Original} & \textbf{Draft} & \textbf{Step} & \textbf{Adaptive}\\
    \hline
    Avg Steps & 111 & 22 & 88 & 50 \\
    Avg Tokens per Step & 51 & 45 & 48 & 55 \\
    \Xhline{1.5pt}
    \end{tabular}
    \caption{Average number of reasoning steps and average tokens per step on MATH500. We use double newlines ("\texttt{\string\n\string\n}") as a step delimiter.}
    \label{tab:step_tokens}
\end{table}
\endgroup




\subsection{Reasoning Behavior Analysis}
\label{sec:reasoning_behavior}

To examine how Draft-Thinking reduces reasoning cost, we compare reasoning traces on MATH500 from the original Qwen3-8B and the three modes of the adapted model. Following ThinkPrune~\cite{hou_thinkprune_2025}, we use DeepSeek-V3-0324 to assign reasoning steps, delimited by double newlines, to ten predefined problem-solving phases.

As shown in Figure~\ref{fig:reasoning_behavior}, draft mode reduces explicit steps across all phases, particularly alternative exploration and repeated verification, while retaining a larger proportion of theorem application and computation. Step mode also reduces steps in most phases, whereas adaptive mode provides an intermediate level of reasoning expansion. Table~\ref{tab:step_tokens} shows that draft mode reduces the average number of steps from 111 to 22, an 80.2\% reduction, while tokens per step decrease only from 51 to 45. Thus, \textbf{most token savings arise from removing redundant reasoning steps rather than compressing the wording of individual steps}.
\section{Conclusion}

We present \textbf{Draft-Thinking}, a lightweight framework for adapting long CoT LRMs to generate efficient, model-adapted draft reasoning. We identify the \textbf{Cross-Scale Draft Mismatch}, where concise traces produced by a strong teacher may omit intermediate transitions required by a smaller target model. To address this mismatch, \textbf{Progressive Draft Adaptation} combines draft-form SFT with tight-budget and hard-task RL to calibrate reasoning granularity to the target model. The adapted model supports draft, adaptive, and step-by-step modes, providing different operating points between efficiency and detailed reasoning. Experiments on Qwen3-8B and Qwen3-4B demonstrate substantial token reductions across mathematical and scientific reasoning benchmarks. In particular,
using only 847 training problems and 120 RL optimization steps, Draft-Thinking offers a low-cost approach to activating efficient draft reasoning while preserving access to long CoT reasoning when needed.

\section*{Limitations}

Draft-Thinking does not compromise the original long-CoT capability and even improves token efficiency across all benchmarks. However, on the most difficult AIME 2025 dataset, the accuracy gap between draft and long-CoT modes is larger than on other sets. We suspect that complex problems require a larger maximum sequence length during draft reasoning training. For instance, on the simpler Math500, draft reasoning already performs nearly as well as long-CoT. Limited by computational resources, we did not further explore the potential of draft reasoning with even longer maximum sequence length.


\bibliography{custom}

@inproceedings{sheng2025verl,
  title={Hybridflow: A flexible and efficient rlhf framework},
  author={Sheng, Guangming and Zhang, Chi and Ye, Zilingfeng and Wu, Xibin and Zhang, Wang and Zhang, Ru and Peng, Yanghua and Lin, Haibin and Wu, Chuan},
  booktitle={Proceedings of the Twentieth European Conference on Computer Systems},
  pages={1279--1297},
  year={2025}
}

@article{miller1956magical,
  title={The magical number seven, plus or minus two: Some limits on our capacity for processing information.},
  author={Miller, George A},
  journal={Psychological review},
  volume={63},
  number={2},
  pages={81},
  year={1956},
  publisher={American Psychological Association}
}

@article{guo2025deepseekr1,
  title={Deepseek-r1: Incentivizing reasoning capability in llms via reinforcement learning},
  author={Guo, Daya and Yang, Dejian and Zhang, Haowei and Song, Junxiao and Zhang, Ruoyu and Xu, Runxin and Zhu, Qihao and Ma, Shirong and Wang, Peiyi and Bi, Xiao and others},
  journal={arXiv preprint arXiv:2501.12948},
  year={2025}
}

@misc{yang2025qwen3technicalreport,
      title={Qwen3 Technical Report}, 
      author={An Yang and Anfeng Li and Baosong Yang and Beichen Zhang and Binyuan Hui and Bo Zheng and Bowen Yu and Chang Gao and Chengen Huang and Chenxu Lv and Chujie Zheng and Dayiheng Liu and Fan Zhou and Fei Huang and Feng Hu and Hao Ge and Haoran Wei and Huan Lin and Jialong Tang and Jian Yang and Jianhong Tu and Jianwei Zhang and Jianxin Yang and Jiaxi Yang and Jing Zhou and Jingren Zhou and Junyang Lin and Kai Dang and Keqin Bao and Kexin Yang and Le Yu and Lianghao Deng and Mei Li and Mingfeng Xue and Mingze Li and Pei Zhang and Peng Wang and Qin Zhu and Rui Men and Ruize Gao and Shixuan Liu and Shuang Luo and Tianhao Li and Tianyi Tang and Wenbiao Yin and Xingzhang Ren and Xinyu Wang and Xinyu Zhang and Xuancheng Ren and Yang Fan and Yang Su and Yichang Zhang and Yinger Zhang and Yu Wan and Yuqiong Liu and Zekun Wang and Zeyu Cui and Zhenru Zhang and Zhipeng Zhou and Zihan Qiu},
      year={2025},
      eprint={2505.09388},
      archivePrefix={arXiv},
      primaryClass={cs.CL},
      url={https://arxiv.org/abs/2505.09388}, 
}

@misc{team_kimi_2025,
	title = {Kimi k1.5: {Scaling} {Reinforcement} {Learning} with {LLMs}},
	shorttitle = {Kimi k1.5},
	url = {http://arxiv.org/abs/2501.12599},
	doi = {10.48550/arXiv.2501.12599},
	urldate = {2025-02-25},
	publisher = {arXiv},
	author = {Team, Kimi and Du, Angang and Gao, Bofei and Xing, Bowei and Jiang, Changjiu and Chen, Cheng and Li, Cheng and Xiao, Chenjun and Du, Chenzhuang and Liao, Chonghua and Tang, Chuning and Wang, Congcong and Zhang, Dehao and Yuan, Enming and Lu, Enzhe and Tang, Fengxiang and Sung, Flood and Wei, Guangda and Lai, Guokun and Guo, Haiqing and Zhu, Han and Ding, Hao and Hu, Hao and Yang, Hao and Zhang, Hao and Yao, Haotian and Zhao, Haotian and Lu, Haoyu and Li, Haoze and Yu, Haozhen and Gao, Hongcheng and Zheng, Huabin and Yuan, Huan and Chen, Jia and Guo, Jianhang and Su, Jianlin and Wang, Jianzhou and Zhao, Jie and Zhang, Jin and Liu, Jingyuan and Yan, Junjie and Wu, Junyan and Shi, Lidong and Ye, Ling and Yu, Longhui and Dong, Mengnan and Zhang, Neo and Ma, Ningchen and Pan, Qiwei and Gong, Qucheng and Liu, Shaowei and Ma, Shengling and Wei, Shupeng and Cao, Sihan and Huang, Siying and Jiang, Tao and Gao, Weihao and Xiong, Weimin and He, Weiran and Huang, Weixiao and Wu, Wenhao and He, Wenyang and Wei, Xianghui and Jia, Xianqing and Wu, Xingzhe and Xu, Xinran and Zu, Xinxing and Zhou, Xinyu and Pan, Xuehai and Charles, Y. and Li, Yang and Hu, Yangyang and Liu, Yangyang and Chen, Yanru and Wang, Yejie and Liu, Yibo and Qin, Yidao and Liu, Yifeng and Yang, Ying and Bao, Yiping and Du, Yulun and Wu, Yuxin and Wang, Yuzhi and Zhou, Zaida and Wang, Zhaoji and Li, Zhaowei and Zhu, Zhen and Zhang, Zheng and Wang, Zhexu and Yang, Zhilin and Huang, Zhiqi and Huang, Zihao and Xu, Ziyao and Yang, Zonghan},
	month = jan,
	year = {2025},
	note = {arXiv:2501.12599 [cs]},
}

@article{hendrycks2021math500,
  title={Measuring mathematical problem solving with the math dataset},
  author={Hendrycks, Dan and Burns, Collin and Kadavath, Saurav and Arora, Akul and Basart, Steven and Tang, Eric and Song, Dawn and Steinhardt, Jacob},
  journal={arXiv preprint arXiv:2103.03874},
  year={2021}
}

@article{lewkowycz2022minervamath,
  title={Solving quantitative reasoning problems with language models},
  author={Lewkowycz, Aitor and Andreassen, Anders and Dohan, David and Dyer, Ethan and Michalewski, Henryk and Ramasesh, Vinay and Slone, Ambrose and Anil, Cem and Schlag, Imanol and Gutman-Solo, Theo and others},
  journal={Advances in neural information processing systems},
  volume={35},
  pages={3843--3857},
  year={2022}
}

@inproceedings{rein2024gpqa,
  title={Gpqa: A graduate-level google-proof q\&a benchmark},
  author={Rein, David and Hou, Betty Li and Stickland, Asa Cooper and Petty, Jackson and Pang, Richard Yuanzhe and Dirani, Julien and Michael, Julian and Bowman, Samuel R},
  booktitle={First Conference on Language Modeling},
  year={2024}
}

@article{he2024olympiadbench,
  title={Olympiadbench: A challenging benchmark for promoting agi with olympiad-level bilingual multimodal scientific problems},
  author={He, Chaoqun and Luo, Renjie and Bai, Yuzhuo and Hu, Shengding and Thai, Zhen Leng and Shen, Junhao and Hu, Jinyi and Han, Xu and Huang, Yujie and Zhang, Yuxiang and others},
  journal={arXiv preprint arXiv:2402.14008},
  year={2024}
}

@article{liu2024livemathbench202505,
  title={Are Your LLMs Capable of Stable Reasoning?},
  author={Liu, Junnan and Liu, Hongwei and Xiao, Linchen and Wang, Ziyi and Liu, Kuikun and Gao, Songyang and Zhang, Wenwei and Zhang, Songyang and Chen, Kai},
  journal={arXiv preprint arXiv:2412.13147},
  year={2024}
}

@misc{xia_tokenskip_2025,
	title = {{TokenSkip}: {Controllable} {Chain}-of-{Thought} {Compression} in {LLMs}},
	shorttitle = {{TokenSkip}},
	url = {http://arxiv.org/abs/2502.12067},
	doi = {10.48550/arXiv.2502.12067},
	
	urldate = {2025-03-12},
	publisher = {arXiv},
	author = {Xia, Heming and Li, Yongqi and Leong, Chak Tou and Wang, Wenjie and Li, Wenjie},
	month = feb,
	year = {2025},
	note = {arXiv:2502.12067 [cs]},
}

@misc{yuan_not_all_tokens_2025,
	title = {Not {All} {Tokens} {Are} {What} {You} {Need} {In} {Thinking}},
	copyright = {Creative Commons Attribution 4.0 International},
	url = {https://arxiv.org/abs/2505.17827},
	doi = {10.48550/ARXIV.2505.17827},
	
	language = {en},
	urldate = {2025-06-19},
	publisher = {arXiv},
	author = {Yuan, Hang and Yu, Bin and Li, Haotian and Yang, Shijun and Wang, Christina Dan and Yu, Zhou and Xu, Xueyin and Qi, Weizhen and Chen, Kai},
	year = {2025},
	note = {Version Number: 1},
}

@misc{cuadron_danger_2025,
	title = {The {Danger} of {Overthinking}: {Examining} the {Reasoning}-{Action} {Dilemma} in {Agentic} {Tasks}},
	shorttitle = {The {Danger} of {Overthinking}},
	url = {http://arxiv.org/abs/2502.08235},
	doi = {10.48550/arXiv.2502.08235},
	urldate = {2025-03-03},
	publisher = {arXiv},
	author = {Cuadron, Alejandro and Li, Dacheng and Ma, Wenjie and Wang, Xingyao and Wang, Yichuan and Zhuang, Siyuan and Liu, Shu and Schroeder, Luis Gaspar and Xia, Tian and Mao, Huanzhi and Thumiger, Nicholas and Desai, Aditya and Stoica, Ion and Klimovic, Ana and Neubig, Graham and Gonzalez, Joseph E.},
	month = feb,
	year = {2025},
	note = {arXiv:2502.08235 [cs]},
	
}

@misc{yang_tops_2025,
	title = {Towards {Thinking}-{Optimal} {Scaling} of {Test}-{Time} {Compute} for {LLM} {Reasoning}},
	url = {http://arxiv.org/abs/2502.18080},
	doi = {10.48550/arXiv.2502.18080},
	urldate = {2025-03-12},
	publisher = {arXiv},
	author = {Yang, Wenkai and Ma, Shuming and Lin, Yankai and Wei, Furu},
	month = feb,
	year = {2025},
	note = {arXiv:2502.18080 [cs]},
}

@misc{chen_do_not_2025,
	title = {Do {NOT} {Think} {That} {Much} for 2+3=? {On} the {Overthinking} of o1-{Like} {LLMs}},
	shorttitle = {Do {NOT} {Think} {That} {Much} for 2+3=?},
	url = {http://arxiv.org/abs/2412.21187},
	doi = {10.48550/arXiv.2412.21187},
	
	urldate = {2025-02-26},
	publisher = {arXiv},
	author = {Chen, Xingyu and Xu, Jiahao and Liang, Tian and He, Zhiwei and Pang, Jianhui and Yu, Dian and Song, Linfeng and Liu, Qiuzhi and Zhou, Mengfei and Zhang, Zhuosheng and Wang, Rui and Tu, Zhaopeng and Mi, Haitao and Yu, Dong},
	month = feb,
	year = {2025},
	note = {arXiv:2412.21187 [cs]},
	
}

@misc{ye_limo_2025,
	title = {{LIMO}: {Less} is {More} for {Reasoning}},
	shorttitle = {{LIMO}},
	url = {http://arxiv.org/abs/2502.03387},
	doi = {10.48550/arXiv.2502.03387},
	urldate = {2025-06-07},
	publisher = {arXiv},
	author = {Ye, Yixin and Huang, Zhen and Xiao, Yang and Chern, Ethan and Xia, Shijie and Liu, Pengfei},
	month = feb,
	year = {2025},
	note = {arXiv:2502.03387 [cs]},
}

@misc{shao_deepseekmath_2024,
	title = {{DeepSeekMath}: {Pushing} the {Limits} of {Mathematical} {Reasoning} in {Open} {Language} {Models}},
	shorttitle = {{DeepSeekMath}},
	url = {http://arxiv.org/abs/2402.03300},
	doi = {10.48550/arXiv.2402.03300},
	
	urldate = {2025-02-19},
	publisher = {arXiv},
	author = {Shao, Zhihong and Wang, Peiyi and Zhu, Qihao and Xu, Runxin and Song, Junxiao and Bi, Xiao and Zhang, Haowei and Zhang, Mingchuan and Li, Y. K. and Wu, Y. and Guo, Daya},
	month = apr,
	year = {2024},
	note = {arXiv:2402.03300 [cs]},
}

@misc{luo_o1-pruner_2025,
	title = {O1-{Pruner}: {Length}-{Harmonizing} {Fine}-{Tuning} for {O1}-{Like} {Reasoning} {Pruning}},
	shorttitle = {O1-{Pruner}},
	url = {http://arxiv.org/abs/2501.12570},
	doi = {10.48550/arXiv.2501.12570},
	urldate = {2025-02-26},
	publisher = {arXiv},
	author = {Luo, Haotian and Shen, Li and He, Haiying and Wang, Yibo and Liu, Shiwei and Li, Wei and Tan, Naiqiang and Cao, Xiaochun and Tao, Dacheng},
	month = jan,
	year = {2025},
	note = {arXiv:2501.12570 [cs]},
}

@misc{kang_c3ot_2024,
	title = {{C3oT}: {Generating} {Shorter} {Chain}-of-{Thought} without {Compromising} {Effectiveness}},
	shorttitle = {{C3oT}},
	url = {http://arxiv.org/abs/2412.11664},
	doi = {10.48550/arXiv.2412.11664},
	urldate = {2025-02-28},
	publisher = {arXiv},
	author = {Kang, Yu and Sun, Xianghui and Chen, Liangyu and Zou, Wei},
	month = dec,
	year = {2024},
	note = {arXiv:2412.11664 [cs]},
}

@misc{han_token-budget-aware_2025,
	title = {Token-{Budget}-{Aware} {LLM} {Reasoning}},
	url = {http://arxiv.org/abs/2412.18547},
	doi = {10.48550/arXiv.2412.18547},
	urldate = {2025-02-28},
	publisher = {arXiv},
	author = {Han, Tingxu and Wang, Zhenting and Fang, Chunrong and Zhao, Shiyu and Ma, Shiqing and Chen, Zhenyu},
	month = feb,
	year = {2025},
	note = {arXiv:2412.18547 [cs]},
}

@misc{song_ConciseR_2025,
	title = {Walk {Before} {You} {Run}! {Concise} {LLM} {Reasoning} via {Reinforcement} {Learning}},
	url = {http://arxiv.org/abs/2505.21178},
	doi = {10.48550/arXiv.2505.21178},
	
	urldate = {2025-06-22},
	publisher = {arXiv},
	author = {Song, Mingyang and Zheng, Mao},
	month = may,
	year = {2025},
	note = {arXiv:2505.21178 [cs]},
}

@misc{cheng_LCR1_2025,
	title = {Optimizing {Length} {Compression} in {Large} {Reasoning} {Models}},
	url = {http://arxiv.org/abs/2506.14755},
	doi = {10.48550/arXiv.2506.14755},
	
	urldate = {2025-07-11},
	publisher = {arXiv},
	author = {Cheng, Zhengxiang and Chen, Dongping and Fu, Mingyang and Zhou, Tianyi},
	month = jun,
	year = {2025},
	note = {arXiv:2506.14755 [cs]},
	
}

@misc{hou_thinkprune_2025,
	title = {{ThinkPrune}: {Pruning} {Long} {Chain}-of-{Thought} of {LLMs} via {Reinforcement} {Learning}},
	shorttitle = {{ThinkPrune}},
	url = {http://arxiv.org/abs/2504.01296},
	doi = {10.48550/arXiv.2504.01296},
	urldate = {2025-07-14},
	publisher = {arXiv},
	author = {Hou, Bairu and Zhang, Yang and Ji, Jiabao and Liu, Yujian and Qian, Kaizhi and Andreas, Jacob and Chang, Shiyu},
	month = apr,
	year = {2025},
	note = {arXiv:2504.01296 [cs]},
}

@misc{xu_chain_of_draft_2025,
	title = {Chain of {Draft}: {Thinking} {Faster} by {Writing} {Less}},
	shorttitle = {Chain of {Draft}},
	url = {http://arxiv.org/abs/2502.18600},
	doi = {10.48550/arXiv.2502.18600},
	urldate = {2025-03-13},
	publisher = {arXiv},
	author = {Xu, Silei and Xie, Wenhao and Zhao, Lingxiao and He, Pengcheng},
	month = mar,
	year = {2025},
	note = {arXiv:2502.18600 [cs]},
}

@misc{aytes_sketch_of_thought_2025,
	title = {Sketch-of-{Thought}: {Efficient} {LLM} {Reasoning} with {Adaptive} {Cognitive}-{Inspired} {Sketching}},
	shorttitle = {Sketch-of-{Thought}},
	url = {http://arxiv.org/abs/2503.05179},
	doi = {10.48550/arXiv.2503.05179},
	urldate = {2025-03-13},
	publisher = {arXiv},
	author = {Aytes, Simon A. and Baek, Jinheon and Hwang, Sung Ju},
	month = mar,
	year = {2025},
	note = {arXiv:2503.05179 [cs]
version: 1},
}

@techreport{openai_gpt5_systemcard_2025,
  title        = {GPT-5 System Card},
  author       = {{OpenAI}},
  institution  = {OpenAI},
  year         = {2025},
  month        = {8},
  url          = {https://cdn.openai.com/gpt-5-system-card.pdf},
  note         = {Accessed: 2025-09}
}

@article{comanici2025gemini,
  title={Gemini 2.5: Pushing the frontier with advanced reasoning, multimodality, long context, and next generation agentic capabilities},
  author={Comanici, Gheorghe and Bieber, Eric and Schaekermann, Mike and Pasupat, Ice and Sachdeva, Noveen and Dhillon, Inderjit and Blistein, Marcel and Ram, Ori and Zhang, Dan and Rosen, Evan and others},
  journal={arXiv preprint arXiv:2507.06261},
  year={2025}
}

\appendix

\section{Experimental Setting Details} 

\subsection{Evaluation Dataset Detail} \label{sec:datasets}
\begin{itemize}

    \item \textbf{MATH500}: 
    A challenging benchmark of 500 high-school competition-level problems spanning seven subjects, including Algebra, Geometry, Number Theory, and Precalculus. Each problem is presented in natural language with LaTeX-formatted notation, providing a robust measure of mathematical reasoning and generalization across diverse topics.
    
    \item \textbf{AIME2025}: 
    A dataset comprising 30 problems from the 2025 American Invitational Mathematics Examination (AIME), a prestigious high-school mathematics competition for top-performing students. Each problem requires deep mathematical insight, multi-step reasoning, and precise problem-solving skills.

    \item \textbf{LiveMathBench}: LiveMathBench is a mathematical dataset specifically designed to include challenging problems from the latest mathematical competitions, thereby avoiding data contamination issues prevalent in existing LLMs and public math benchmarks. We use the 202505 version of LiveMathBench, which contains 100 mathematical questions from various countries, including non-English problems.

    \item \textbf{OlympiadBench}: A subset of the original OlympiadBench dataset containing 675 problems from Olympiad-level mathematics and physics competitions, including the Chinese college entrance examination.
    
    \item \textbf{MinervaMath}: An undergraduate-level dataset containing 272 problems in physics, biology, chemistry, economics, and other sciences that require quantitative reasoning.
    
    \item \textbf{GPQA-Diamond}: A challenging dataset consisting of 198 graduate-level multiple-choice questions written by domain experts across biology, physics, and chemistry.
\end{itemize}

\subsection{Prompt Templates}
\label{appendix:prompts}
Table~\ref{tab:prompts} shows the detailed designs of three prompts: $\boldsymbol{p}_{\text{step}}$ for long CoT reasoning, $\boldsymbol{p}_{\text{draft}}$ for draft reasoning, and $\boldsymbol{p}_{\text{adaptive}}$ for instance-adaptive reasoning.

\begin{table}[t]
\footnotesize
    \centering
    \renewcommand{\arraystretch}{1.2}
    \resizebox{0.98\linewidth}{!}{
    \begin{tabular}{c p{0.65\linewidth}}
    \Xhline{1.5pt}
    \rowcolor[HTML]{C0C0C0}
     \textbf{Prompt Method} & \textbf{Content} \\
     \hline
     \rowcolor[HTML]{F5F5F5}
    Long CoT  & Please reason step by step, and put your final answer within \textbackslash boxed\{\}. \\
    \rowcolor[HTML]{E5E4E2}
    Draft  & Let's think step by step, but only keep a minimum draft for each thinking step, with as few words as possible, and output the final answer within \textbackslash boxed\{\}. \\
    \rowcolor[HTML]{DBD7D2}
    Instance adaptive & First, you should decide the thinking mode based on the question's difficulty. The first mode is the normal way, where you can think in detail. The second mode is to keep only a minimal draft with as few words as possible. Then reason step by step, and output the final answer within \textbackslash boxed\{\}. \\
    \Xhline{1.5pt}
    \end{tabular}
    }
    \caption{Detailed designs of three prompts: $\boldsymbol{p}_{\text{step}}$ for long CoT reasoning, $\boldsymbol{p}_{\text{draft}}$ for draft reasoning, and $\boldsymbol{p}_{\text{adaptive}}$ for instance-adaptive reasoning.}
    \label{tab:prompts}
\end{table}

\begin{figure*}[t]
  \includegraphics[width=1\textwidth]{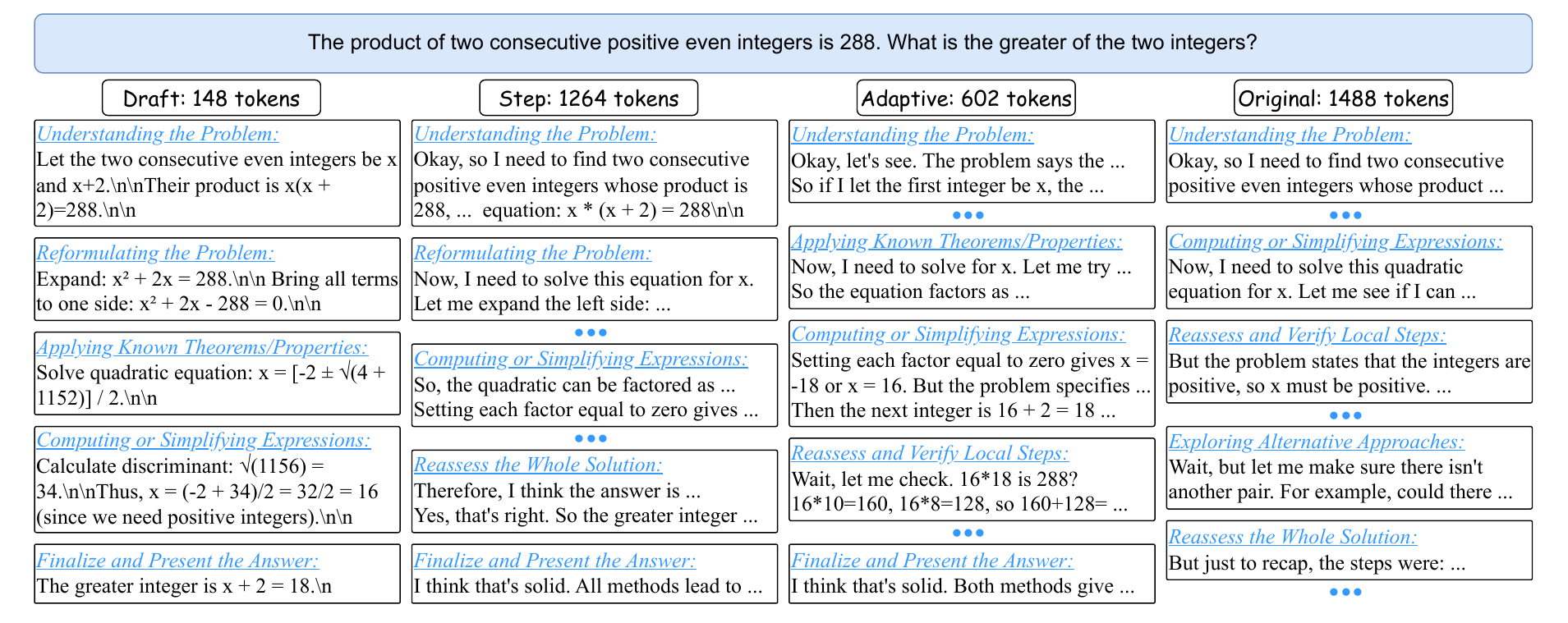}
  \caption{Example resoning trace of original Qwen3-8B and Draft thinking on a MATH500 question.}
  \label{fig:case}
\end{figure*}


\subsection{Evaluation protocol Detail.} \label{appendix:evaluation} 
We report pass@1 accuracy (\textbf{ACC}), the average number of generated response tokens (\textbf{Tokens}), and token efficiency (\textbf{EFF}).

Following the evaluation protocol in LIMO~\cite{ye_limo_2025}, we use greedy decoding with a single sample for MATH500, MinervaMath, OlympiadBench, and GPQA-Diamond.
For the smaller AIME2025 and LiveMathBench datasets, we generate 16 and 6 samples per problem, respectively, with temperature $0.6$ and top-$p$ $0.95$, and compute unbiased pass@1. The maximum generation length is set to 32,768 tokens for all methods, and the same answer-extraction evaluator based on Qwen-2.5-Math \cite{guo2025deepseekr1} is used throughout.

Token efficiency is defined as
\begin{equation}
    \mathrm{EFF}
    =
    \frac{\mathrm{ACC}}{\mathrm{Tokens}}
    \times 100,
\end{equation}
where ACC is expressed in percentage points.
EFF summarizes the amount of accuracy obtained per 100 generated tokens.
We interpret it jointly with ACC and Tokens rather than as a standalone performance measure.

\subsection{Controlled baselines.}
\label{appendix:baselines}
\textbf{SFT-only} fine-tunes Qwen3-8B on original LIMO~\cite{ye_limo_2025} dataset.

\textbf{GRPO-only}~\cite{shao_deepseekmath_2024} trains the model with GRPO on a pooled mixture of LIMO and AIME2024, using a maximum response length of
$L_{\max}=6{,}000$.

\textbf{ThinkPrune}~\cite{hou_thinkprune_2025} reproduces the two-stage RL procedure using the same pooled LIMO and AIME2024 training data, with
$L_{\max}=6{,}000$ and $3{,}000$ in the first and second stages, respectively.

Unlike Draft-Thinking, both RL baselines mix the LIMO and AIME2024 examples throughout training rather than introducing them progressively.
To control for the policy-optimization budget, GRPO-only, ThinkPrune, and Draft-Thinking are each trained for a total of 120 RL optimization steps.

\subsection{Implementation Details}
\label{appendix:imple details}

We utilize the Verl framework~\cite{sheng2025verl} for both supervised fine-tuning and reinforcement learning. All RL experiments are conducted
on 6 NVIDIA L20-48G GPUs. Tables~\ref{tab:verl_settings_8B} and
\ref{tab:verl_settings_4B} summarize the main RL training configurations
of Draft-Thinking on Qwen3-8B and Qwen3-4B, respectively.

\paragraph{RL Training.}
The two RL stages of Draft-Thinking use the same optimization framework
with different training data and maximum response lengths. Specifically,
D-RL3k is trained on $D_{\mathrm{rl}}$ with a maximum response length of
3,000 tokens, while D-RL6k is trained on AIME2024 with a maximum response
length of 6,000 tokens. Unless otherwise specified, all experiments,
including baselines (GRPO-only, ThinkPrune) and ablation variants, follow the same training
configuration under the corresponding maximum response length to ensure
a fair comparison. For baseline-specific methods, we only modify the
components required by their original training procedures. For example,
GRPO-only uses the same 6k response length and optimization settings as
D-RL6k, except for the training data and the number of epochs (30 epochs,
corresponding to 120 optimization steps).

\begingroup
\setlength{\tabcolsep}{3pt}
\begin{table}[t]
\small
    \centering
    \begin{tabular}{lcc}
    \Xhline{1.5pt}
    \rowcolor[HTML]{D3D3D3}
    \textbf{Parameter} & \textbf{D-RL3k} & \textbf{D-RL6k} \\
    \hline
    datasets & $D_{\text{rl}}$ & AIME2024 \\
    advantage estimator & \multicolumn{2}{c}{grpo} \\
    train batch size & 96 & 30\\
    max prompt length & 240 & 500 \\
    max response length & 3000 & 6000 \\
    actor optim lr & \multicolumn{2}{c}{1e-6} \\
    use remove padding& \multicolumn{2}{c}{True} \\
    ppo mini batch size & 48 & 30 \\
    ppo micro batch size per gpu & 2 & 1 \\
    use kl loss & \multicolumn{2}{c}{False} \\
    entropy coeff & \multicolumn{2}{c}{0} \\
    enable gradient checkpointing & \multicolumn{2}{c}{True} \\
    model parallel size & \multicolumn{2}{c}{2} \\
    actor rollout engine & \multicolumn{2}{c}{vllm}\\
    rollout num per question & \multicolumn{2}{c}{6} \\
    gpu memory utilization & 0.6& 0.5 \\
    use kl in reward & \multicolumn{2}{c}{False} \\
    num nodes & \multicolumn{2}{c}{1} \\
    gpus per node & \multicolumn{2}{c}{6 (L20-48G)} \\
    total epochs & 15& 60\\
    \Xhline{1.5pt}
    \end{tabular}
    \caption{Main RL training configurations of Draft-Thinking on Qwen3-8B using the Verl framework. D-RL3k and D-RL6k denote the two progressive RL stages with maximum response lengths of 3,000 and 6,000 tokens, respectively.}
    \label{tab:verl_settings_8B}
\end{table}
\endgroup

\begingroup
\setlength{\tabcolsep}{3pt}
\begin{table}[t]
\small
    \centering
    \begin{tabular}{lcc}
    \Xhline{1.5pt}
    \rowcolor[HTML]{D3D3D3}
    \textbf{Parameter} & \textbf{D-RL3k} & \textbf{D-RL6k} \\
    \hline
    datasets & $D_{\text{rl}}$ & AIME2024 \\
    advantage estimator & \multicolumn{2}{c}{grpo} \\
    train batch size & 96 & 30\\
    max prompt length & 240 & 500 \\
    max response length & 3000 & 6000 \\
    actor optim lr & \multicolumn{2}{c}{1e-6} \\
    use remove padding& \multicolumn{2}{c}{True} \\
    ppo mini batch size & 48 & 30 \\
    ppo micro batch size per gpu & \multicolumn{2}{c}{2} \\
    use kl loss & \multicolumn{2}{c}{False} \\
    entropy coeff & \multicolumn{2}{c}{0} \\
    enable gradient checkpointing & \multicolumn{2}{c}{True} \\
    model parallel size & 1 & 2 \\
    actor rollout engine & \multicolumn{2}{c}{vllm}\\
    rollout num per question & \multicolumn{2}{c}{6} \\
    gpu memory utilization & \multicolumn{2}{c}{0.6} \\
    use kl in reward & \multicolumn{2}{c}{False} \\
    num nodes & \multicolumn{2}{c}{1} \\
    gpus per node & \multicolumn{2}{c}{6 (L20-48G)} \\
    total epochs & 15& 60\\
    \Xhline{1.5pt}
    \end{tabular}
    \caption{Main RL training configurations of Draft-Thinking on Qwen3-4B using the Verl framework. D-RL3k and D-RL6k denote the two progressive RL stages with maximum response lengths of 3,000 and 6,000 tokens, respectively. }
    \label{tab:verl_settings_4B}
\end{table}
\endgroup

\begingroup
\begin{table*}[t]
\centering
\begin{tabular}{lccc}
\Xhline{1.5pt}
\rowcolor[HTML]{D3D3D3}
\textbf{Parameter} 
& \textbf{SFT-only}
& \textbf{Draft-SFT (8B)}
& \textbf{Draft-SFT (4B)}
\\
\hline
Base model 
& Qwen3-8B 
& Qwen3-8B 
& Qwen3-4B
\\
Training data 
& LIMO
& $D_{\text{rl}}$
& $D_{\text{rl}}$
\\
Training samples 
& 817
& 342
& 342
\\
Max sequence length 
& 10,000
& 6,000
& 6,000
\\
Global batch size 
& 60
& 60
& 60
\\
Micro batch size/GPU
& 1
& 1
& 2
\\
Learning rate
& 5e-6
& 5e-6
& 5e-6
\\
Precision
& BF16
& BF16
& BF16
\\
Training epochs
& 15
& 15
& 15
\\
GPUs
& 6$\times$L20-48G
& 6$\times$L20-48G
& 6$\times$L20-48G
\\
\Xhline{1.5pt}
\end{tabular}
\caption{
SFT training configurations. Draft-SFT uses the filtered
DeepSeek-V3-generated draft reasoning dataset $D_{\text{rl}}$, while SFT-only uses the original LIMO solutions. The maximum sequence length refers to
the combined prompt and response sequence.
}
\label{tab:sft_settings}
\end{table*}
\endgroup

\paragraph{Supervised fine-tuning.}
We perform supervised fine-tuning using the Verl framework.
Table~\ref{tab:sft_settings} summarizes the SFT configurations.
The SFT-only baseline is trained on the original LIMO solutions,
whereas Draft-SFT uses 342 correct draft-style solutions generated by
DeepSeek-V3 after filtering incorrect generations.
All SFT experiments use BF16 precision and the AdamW optimizer with a
learning rate of $5\times10^{-6}$ for 15 epochs.

\subsection{Chunked Symbolism prompt\cite{aytes_sketch_of_thought_2025}} \label{sec:chunked_symbolism}
\begin{figure*}[t]
  \includegraphics[width=1\textwidth]{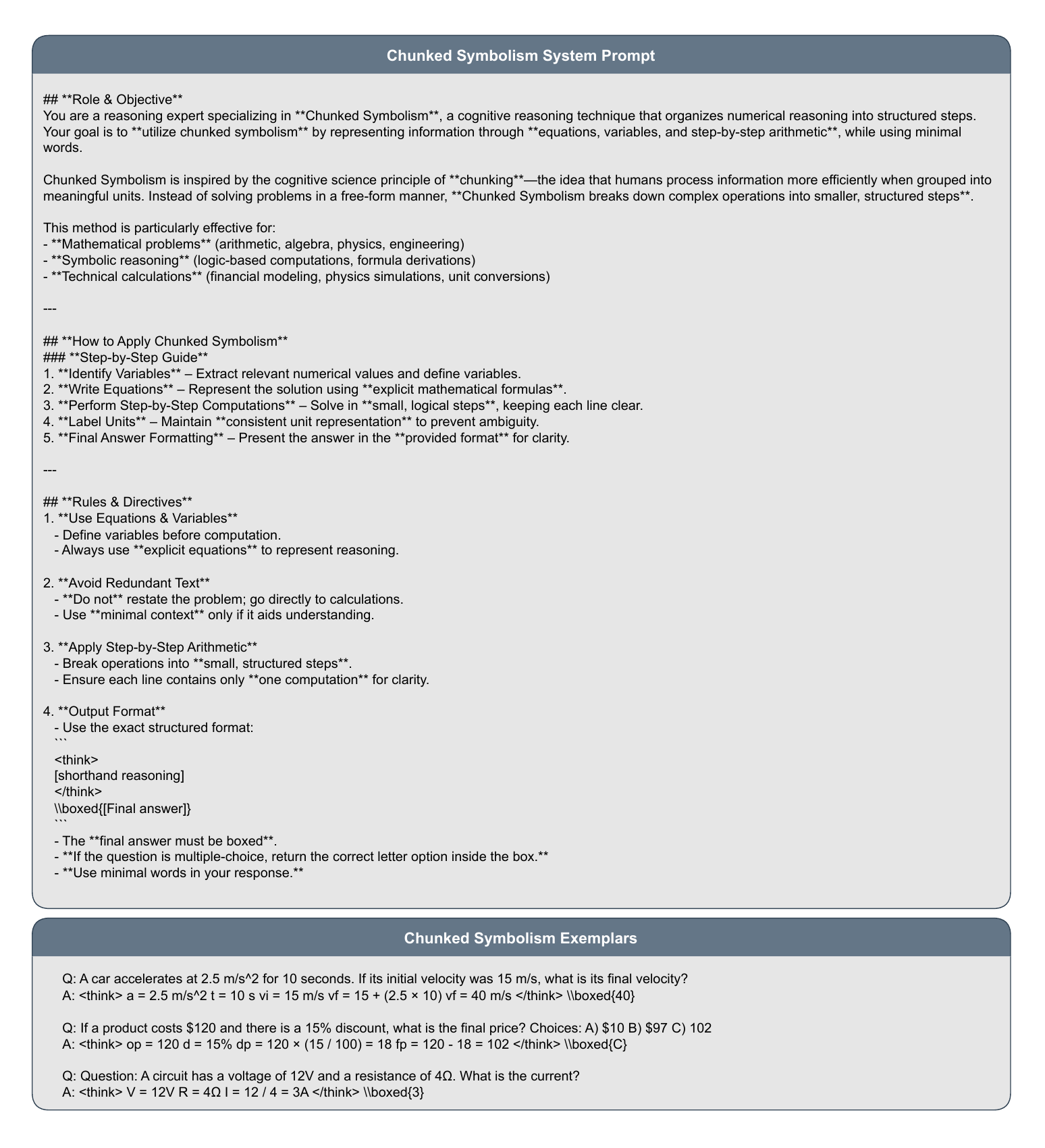}
  \caption{Chunked Symbolism Prompt \cite{aytes_sketch_of_thought_2025}.}
  \label{fig:chunked_symbolism}
\end{figure*}

Chunked symbolism is based on working memory chunking theory \cite{miller1956magical}, which condenses mathematical reasoning into dense symbolic representations containing more information with fewer tokens. The detailed prompt of Chunked Symbolism is shown in Figure~\ref{fig:chunked_symbolism}.

\subsection{Reasoning behavior analysis prompt \cite{hou_thinkprune_2025}}
\begin{figure*}[t]
  \includegraphics[width=1\textwidth]{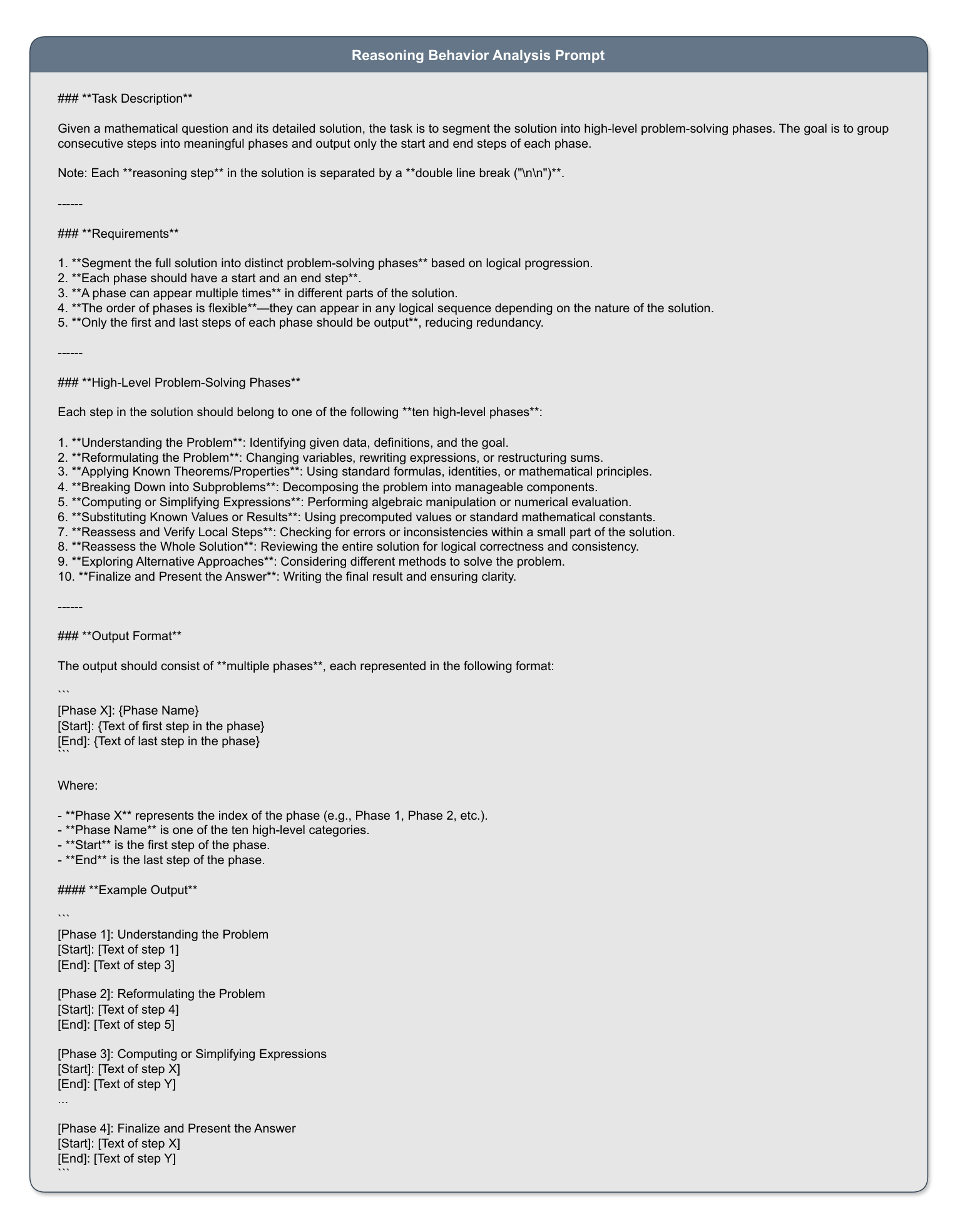}
  \caption{Reasoning behavior analysis prompt \cite{hou_thinkprune_2025}.}
  \label{fig:reasoning_behavior_analysis_prompt}
\end{figure*}

Figure~\ref{fig:reasoning_behavior_analysis_prompt} shows the detail prompt for reasoning behavior analysis.

\section{Additional Experiments}

\subsection{Case Study}

Figure~\ref{fig:case} presents a detailed case comparison between the original Qwen3-8B and Draft-Thinking on a MATH500 sample.

\subsection{Training Pipeline Analysis}
\label{appendix:training pipeline}

\begingroup
\setlength{\tabcolsep}{3pt}
\begin{table*}[t]
\small
  \centering
  \begin{tabular}{lccccccccccccc}
    \Xhline{1.5pt}
    \multirow[m]{2}{*}{\textbf{Method}} & \multirow[m]{2}{*}{\textbf{Prompt}} & \multicolumn{3}{c}{\textbf{Minerva}} & \multicolumn{3}{c}{\textbf{AIME2025}} & \multicolumn{3}{c}{\textbf{LiveMathBench}} & \multicolumn{3}{c}{\textbf{MATH500}}\\
    & &  Acc & Tokens & EFF & Acc & Tokens & EFF & Acc & Tokens & EFF & Acc & Tokens & EFF\\
    \hline
    \multicolumn{14}{c}{\texttt{Qwen3-4B}} \\
    \hline
    \rowcolor[HTML]{D3D3D3}
    Original & step & 51.47 & 7253 & 0.71 & 63.12 & 17703 & 0.36 & 29 & 16813 & 0.17 & 92.2 & 5505 & 1.67 \\
    \hline
    \multirow[t]{3}{*}{D-SFT->D-RL3k->D-RL6k} & draft & 46.69 & 1421 & \underline{3.28} & 46.04 & 7097 & \textbf{0.65} & 22.67 & 6051 & \textbf{0.38} & 92 & 1269 & \textbf{7.24} \\
     & step & 48.9 & 4815 & 1.01 & 60.21	& 12885 & 0.46 & 26.67 & 12863 & 0.21 & 92.8 & 4130 & 2.24\\
     & adaptive & 48.9 & 3620 & 1.35 & 53.33 & 10048 & 0.53 & 24.17 & 8353 & 0.29 & 92.8 & 2292 & 4.04\\
    \hline  
    \multirow[t]{3}{*}{D-RL3k->D-RL6k} & draft & 49.26 & 1669& 2.95 & 44.79 & 7720 & \underline{0.58} & 21.17 & 6242& \underline{0.34} & 92.4 &1531& 6.03 \\
    & step & 51.84 & 3716 & 1.39 & 51.67 & 12186 & 0.42 & 27.83 & 11626 & 0.24& 92.8 & 3309 & 2.80 \\
    & adaptive & 51.84 & 2206 & 2.34 & 49.17 & 10304 & 0.47 & 23.17 & 8751 & 0.26 & 92 & 2342 & 3.92\\
    \hline 
    \multirow[t]{3}{*}{D-RL6k->D-RL3k} & draft & 50 & 1323 & \textbf{3.77} & 45 & 7787 & 0.57 & 19.17 & 6292& 0.30 & 92.2& 1525& \underline{6.04} \\
    & step & 49.63 & 3936 & 1.26 & 54.37 & 11731 & 0.46 & 26.5 &11626& 0.23  & 93.2 & 2952 & 3.15 \\
    & adaptive &50 & 2392 & 2.08 & 48.13 & 9903 & 0.48 & 26.33 & 8358&  0.31& 91.4 & 2262 & 4.04 \\
    \hline 
    RL6k->RL3k & step & 51.47 & 2785 & 1.84 & 47.71 & 10951 & 0.43 & 21.83 & 9528&  0.23 & 92& 2598& 3.54 \\
    \Xhline{1.5pt}
  \end{tabular}
  \caption{\label{tab:ablation}
    Comparison of different training strategies on Qwen3-4B. 
  }
\end{table*}
\endgroup

\begin{figure}[t]
  \includegraphics[width=1\columnwidth]{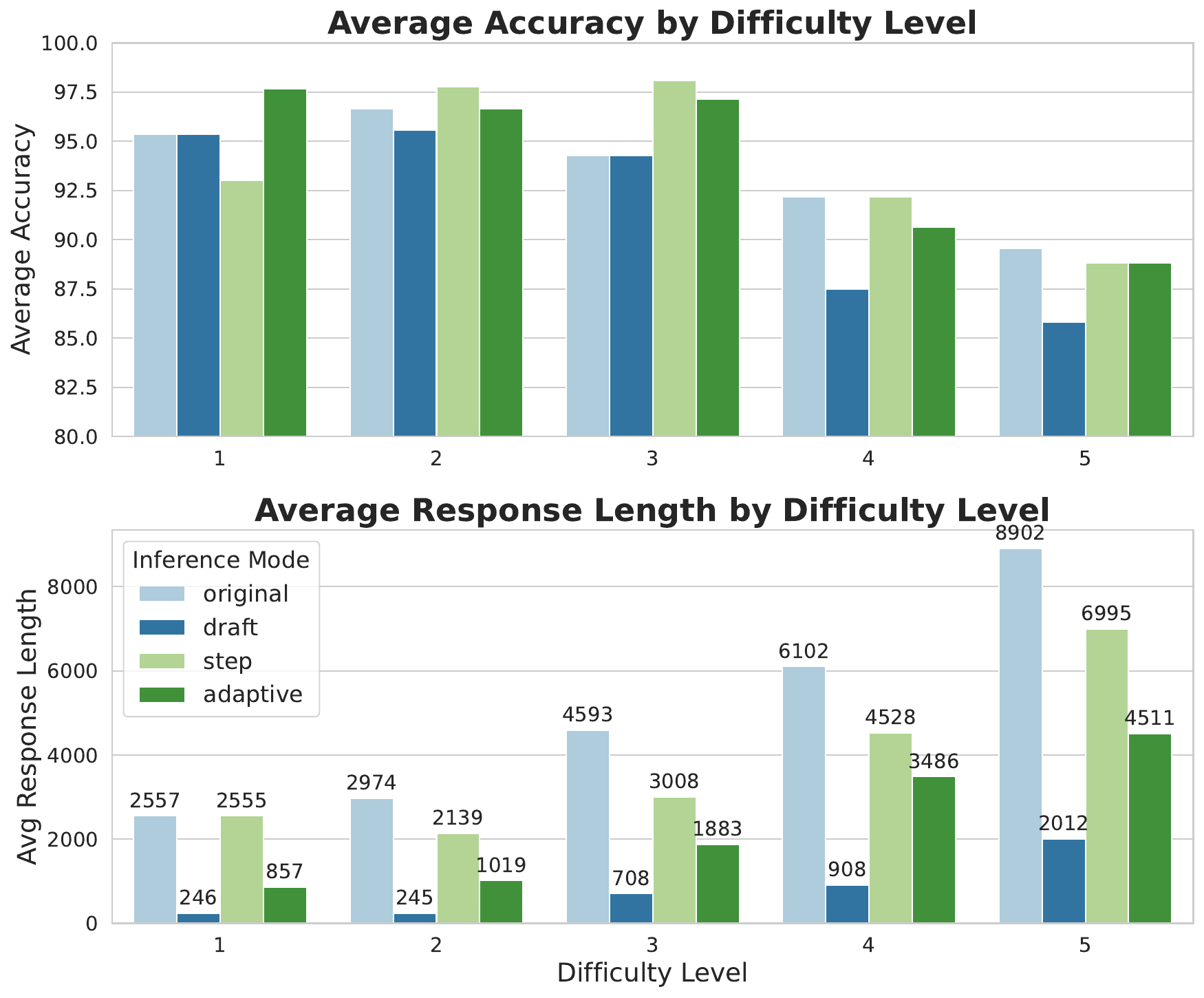}
  \caption{Comparison of average accuracy and response length across different difficulty levels on MATH500.}
  \label{fig:level_math500}
\end{figure}
\begin{figure}[t]
  \includegraphics[width=1\columnwidth]{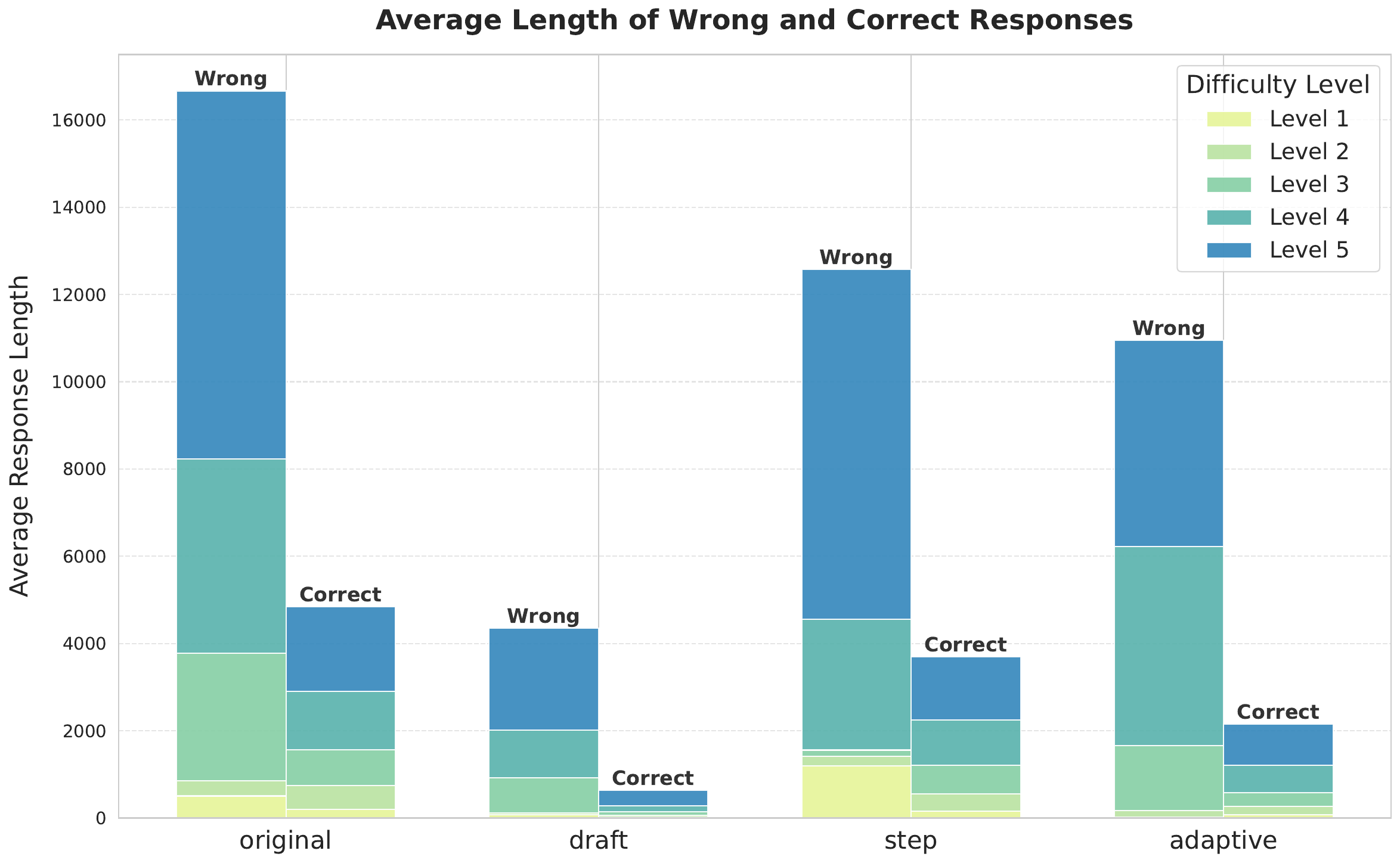}
  \caption{Average lengths of correct vs. wrong responses on MATH500.}
  \label{fig:wrong_correct}
\end{figure}

To further investigate the contribution of each component in
Progressive Draft Adaptation, we conduct additional experiments on
Qwen3-4B with several training variants.
Specifically, we compare:
(1) \textbf{D-RL3k$\rightarrow$D-RL6k}, which removes draft-form
initialization and directly performs two-stage RL under the draft
prompt;
(2) \textbf{ThinkPrune (RL6k$\rightarrow$RL3k)}, which applies
iterative length pruning with the step-by-step prompt; and
(3) \textbf{D-RL6k$\rightarrow$D-RL3k}, which introduces the draft
prompt into the same progressive pruning strategy.
All variants use identical training data for the corresponding RL
stages, namely $D_{\mathrm{rl}}$ and AIME2024 for the 3k and 6k
settings, respectively.


Table~\ref{tab:ablation} shows that the complete Draft-Thinking pipeline achieves the strongest overall accuracy-efficiency trade-off. Removing draft-form initialization (D-RL3k$\rightarrow$D-RL6k) consistently reduces draft-mode EFF compared with the complete method, demonstrating that Draft SFT provides an effective concise reasoning initialization. Moreover, on AIME2025, its step-mode accuracy is 8.54 percentage points lower than Draft-Thinking
(60.21\% vs. 51.67\%), indicating that 
\textbf{draft-form initialization elevates the performance ceiling of draft reasoning EFF and plays a crucial role in enabling mode discrimination, thereby preserving long CoT reasoning quality}. 

Comparing the two length-adaptation strategies, D-RL6k$\rightarrow$D-RL3k
outperforms ThinkPrune under comparable settings, achieving higher draft-mode EFF while maintaining similar accuracy. This suggests that optimizing RL under the draft prompt is more effective than
directly pruning long-CoT trajectories. In addition, although D-RL3k$\rightarrow$D-RL6k and D-RL6k$\rightarrow$D-RL3k achieve comparable efficiency without draft-form initialization, introducing Draft SFT
substantially improves the performance ceiling of the D-RL3k$\rightarrow$D-RL6k strategy, leading to the final Draft-Thinking model with stronger overall efficiency.

Overall, the Qwen3-4B experiments demonstrate that Draft-Thinking remains effective under a smaller model scale. Despite the reduced model capacity, the complete pipeline consistently achieves favorable accuracy-efficiency trade-offs across multiple benchmarks and supports
the same draft, step-by-step, and adaptive reasoning modes. These results indicate that Progressive Draft Adaptation is not limited to a specific model size, but provides a general approach for activating efficient reasoning behaviors in existing long CoT models.

\begin{figure*}[t]
  \includegraphics[width=1\textwidth]{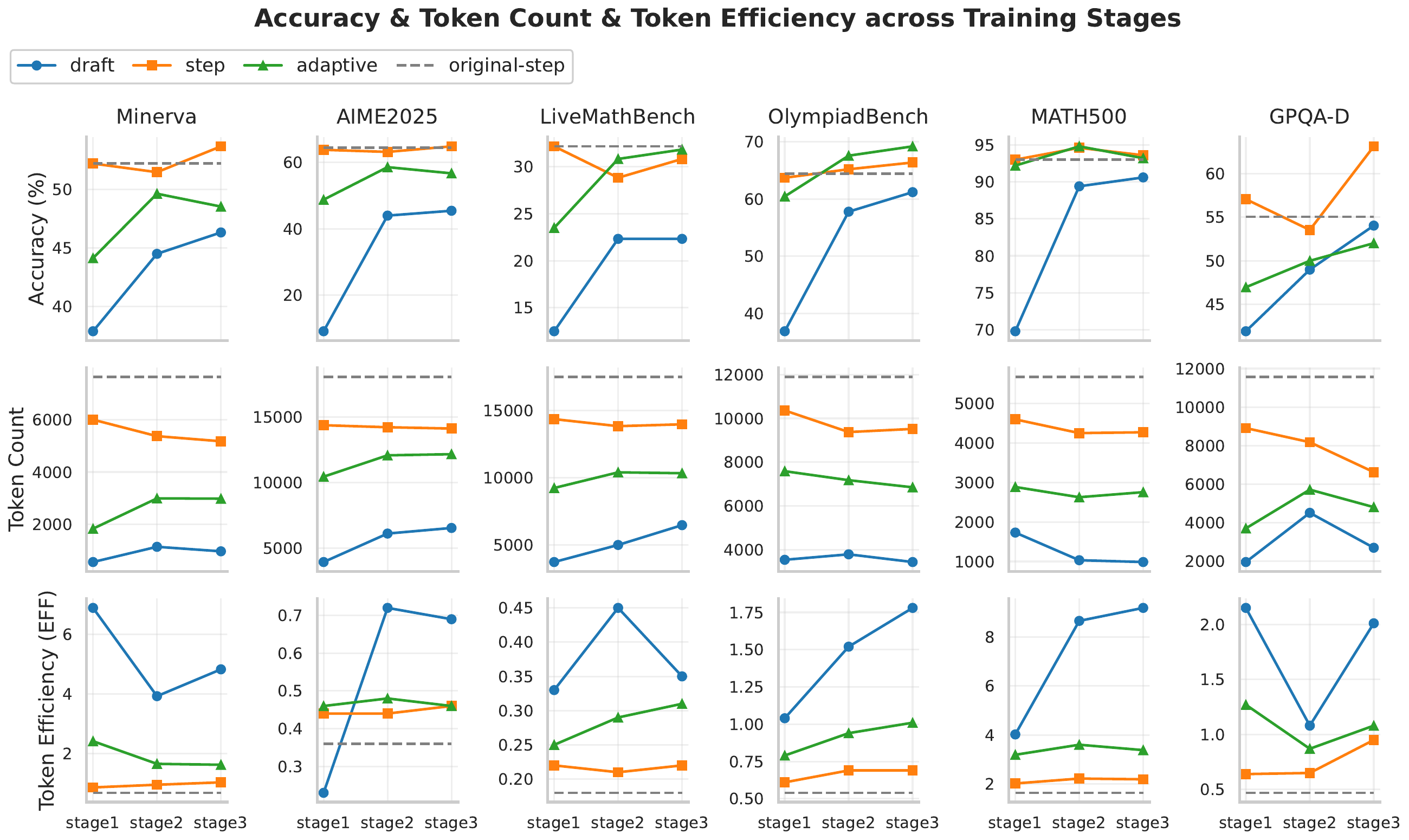}
  \caption{Accuracy, token count, and token efficiency trends for three reasoning modes (draft, long CoT, and adaptive) across the three training stages of Draft-Thinking on Qwen3-8B.}
  \label{fig:acc_eff_stages}
\end{figure*}

\begin{figure*}[t]
  \includegraphics[width=1\textwidth]{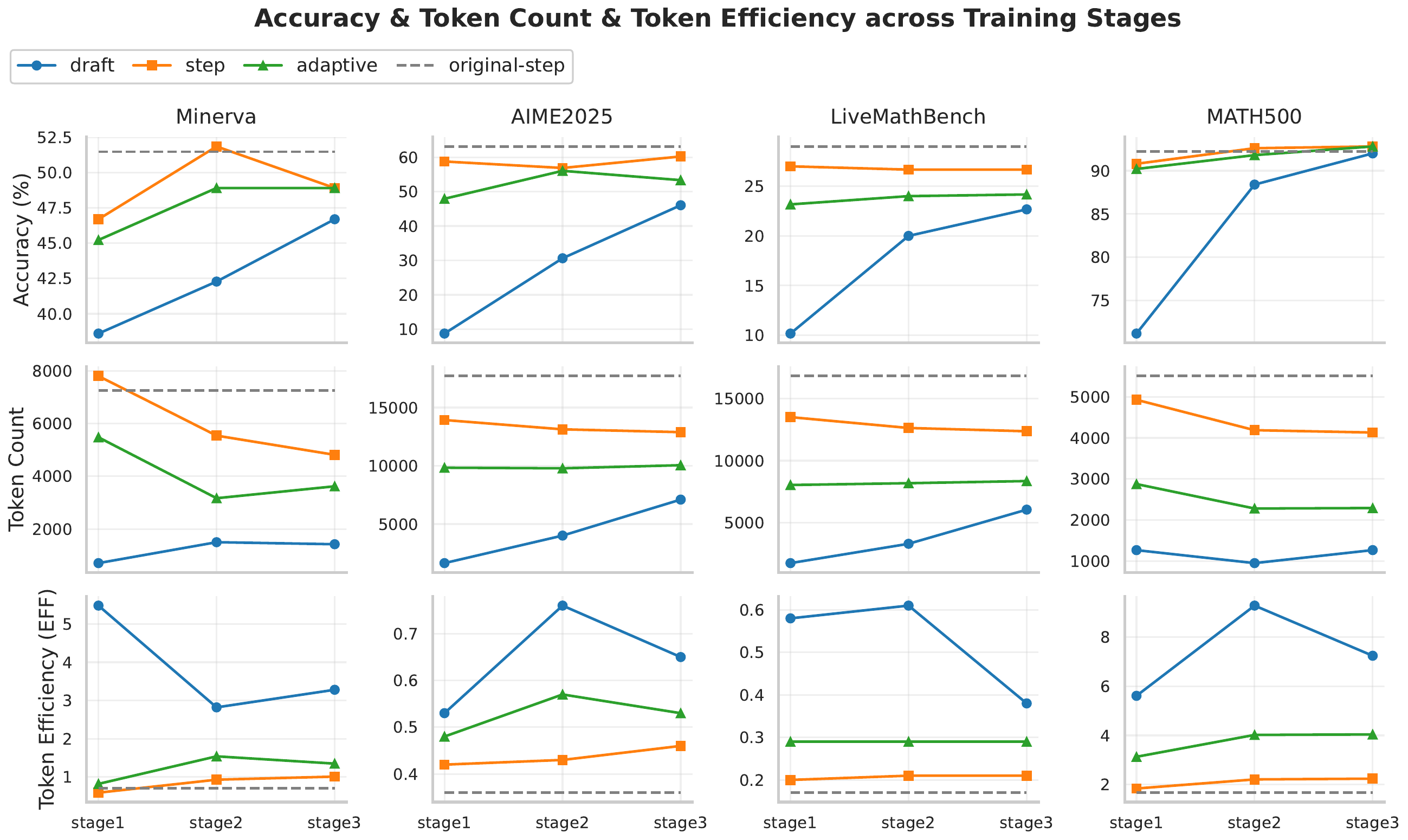}
  \caption{Accuracy, token count, and token efficiency trends for three reasoning modes (draft, long CoT, and adaptive) across the three training stages of Draft Thinking on Qwen3-4B.}
  \label{fig:acc_eff_stages_4B}
\end{figure*}

\subsection{Difficulty-Level Analysis }

As illustrated in Figure~\ref{fig:level_math500}, on Qwen3-8B, draft reasoning maintains comparable accuracy to the original model on low- and medium-difficulty problems, with accuracy degradation mainly occurring on high-difficulty levels (Levels 4 and 5). In contrast, adaptive reasoning and step-by-step reasoning achieve comparable performance across different difficulty levels. Notably, the largest token reductions are observed on Level 1--3 problems; for example, the average response length is reduced to approximately one-tenth of the original model on Levels 1--2 problems.

Figure~\ref{fig:wrong_correct} further reveals a substantial difference
between correct and incorrect responses: incorrect responses tend to be significantly longer than correct ones across different reasoning modes.
This suggests that longer reasoning traces do not necessarily guarantee
correctness, while improving draft reasoning on high-difficulty tasks
may require a larger reasoning budget to recover the missing intermediate
transitions.

\subsection{Training Dynamics across Draft-Thinking Stages}


Figures~\ref{fig:acc_eff_stages} and~\ref{fig:acc_eff_stages_4B} illustrate the evolution of accuracy, token count, and token efficiency across the three Draft-Thinking training stages on Qwen3-8B and Qwen3-4B. We observe three consistent patterns.

First, Draft-Thinking training largely preserves the original long-CoT reasoning capability. As shown by the step-by-step mode, accuracy remains comparable to or improves over the original model across most benchmarks, while the generated responses become substantially shorter.

Second, progressive RL effectively enhances draft reasoning quality. Across most benchmarks, the draft mode gains accuracy after tight-budget adaptation while maintaining a compact reasoning budget. For easier and medium-difficulty tasks, the hard-task adaptation stage can further reduce response length while improving accuracy, whereas harder benchmarks such as AIME2025 and LiveMathBench require longer reasoning traces to recover additional intermediate transitions.

Third, improvements in draft reasoning efficiency are transferred to other inference modes. The step-by-step and adaptive modes also achieve better token efficiency after Draft-Thinking training, demonstrating that learning an efficient draft behavior improves the overall
reasoning efficiency of the adapted model.

\end{document}